\documentclass[preprint,12pt,authoryear]{elsarticle}

\usepackage{amssymb}
\usepackage{makecell}
\usepackage{booktabs}
\usepackage{graphicx}
\usepackage{multirow}
\usepackage{float}
\usepackage{subfig}
\usepackage{color}
\usepackage{amsmath,bm}
\usepackage{changes}

\usepackage[export]{adjustbox} 
\graphicspath{{./figure/}}

\usepackage{lineno}

\usepackage{hyperref}

\usepackage{color, soul}
\setulcolor{red}
\setstcolor{blue}
\sethlcolor{yellow}

\begin{document}
\begin{sloppypar}

\begin{frontmatter}



\title{Aerial-Ground Image Feature Matching via 3D Gaussian Splatting-based Intermediate View Rendering}


\author[inst1]{Jiangxue Yu}
\author[inst1]{Hui Wang}

\affiliation[inst1]{organization={School of Computer Science},
            addressline={China University of Geosciences}, 
            city={Wuhan},
            postcode={430074}, 
            country={China}}

\author[inst2,inst3,inst4]{San Jiang\corref{corresponding}}
\ead{jiangsan@szu.edu.cn}
\cortext[corresponding]{Corresponding author.}

\author[inst2,inst3]{Xing Zhang}
\author[inst2,inst3]{Dejin Zhang}
\author[inst2,inst3]{Qingquan Li}

\affiliation[inst2]{organization={Guangdong Key Laboratory of Urban Informatics},
            addressline={Shenzhen University}, 
            city={Guangdong Shenzhen},
            postcode={518060}, 
            country={China}}

\affiliation[inst3]{organization={MNR Key Laboratory for Geo-Environmental Monitoring of Great Bay Area},
            addressline={Shenzhen University}, 
            city={Guangdong Shenzhen},
            postcode={518060}, 
            country={China}}

\affiliation[inst4]{organization={Engineering Research Center of Natural Resource Information Management and Digital Twin Engineering Software},
            addressline={Ministry of Education}, 
            city={Wuhan},
            postcode={430074}, 
            country={China}}
            
\begin{abstract}
The integration of aerial and ground images has been a promising solution in 3D modeling of complex scenes, which is seriously restricted by finding reliable correspondences. The primary contribution of this study is a feature matching algorithm for aerial and ground images, whose core idea is to generate intermediate views to alleviate perspective distortions caused by the extensive viewpoint changes. First, by using aerial images only, sparse models are reconstructed through an incremental SfM (Structure from Motion) engine due to their large scene coverage. Second, 3D Gaussian Splatting is then adopted for scene rendering by taking as inputs sparse points and oriented images. For accurate view rendering, a render viewpoint determination algorithm is designed by using the oriented camera poses of aerial images, which is used to generate high-quality intermediate images that can bridge the gap between aerial and ground images. Third, with the aid of intermediate images, reliable feature matching is conducted for match pairs from render-aerial and render-ground images, and final matches can be generated by transmitting correspondences through intermediate views. By using real aerial and ground datasets, the validation of the proposed solution has been verified in terms of feature matching and scene rendering and compared comprehensively with widely used methods. The experimental results demonstrate that the proposed solution can provide reliable feature matches for aerial and ground images with an obvious increase in the number of initial and refined matches, and it can provide enough matches to achieve accurate ISfM reconstruction and complete 3DGS-based scene rendering.
\end{abstract}




\end{frontmatter}

\section{Introduction}
\label{sec:1}
3D reconstruction has been extensively explored within the domain of photogrammetry and remote sensing (RS), and it plays a key role in urban scene modeling and management \citep{ge2024rapid,wang2023oblique}. In recent years, low-altitude RS platforms, such as unmanned aerial vehicle (UAV), can provide images with high spatial and time resolutions for the detailed 3D modeling of urban scenes, except for satellite images that aim at covering large-scale scenes. Integrated with geometry-aware trajectory planning techniques, such as the optimized views photogrammetry and snap-to-the-object photogrammetry, UAV can achieve flexible recording of ground objects and collect multi-view high-resolution images \citep{li2023optimized}. However, 3D modeling in urban canyons still faces great challenges because of the limited observation space of aerial RS platforms and serious occlusions of urban high buildings. Compared with aerial RS platforms, ground mobile mapping systems (MMS) have the advantage of recording near-ground scenes, which could be a well compensation to provide complete observation \citep{jhan2021integrating}. Thus, the integration of aerial and ground images has been a promising solution in 3D modeling of complex urban scenes \citep{gao2019ground}.

The integration of aerial and ground images mainly relies on their reliable registration. In the literature, there are two major groups of solutions to achieve this purpose \citep{gao2019ground}. For the first one, by using the POS (Positioning and Orientation System) data from RS platforms, sparse 3D models can be separately created from aerial and ground images and roughly geo-referenced to the same coordinate system. The integration of aerial and ground images is then converted as the registration of point clouds of two sparse models \citep{gao2018accurate,gao2018ancient,shan2014accurate}, which can be achieved by using a reliable geometric transformation estimation algorithm, such as ICP (Iterative Closest Point) \citep{besl1992method}. However, the registration accuracy of the fused model would be seriously degenerated due to the limited overlap and existing outliers in the sparse models. For the second group, feature matching is further conducted to establish correspondences between aerial and ground images, which are then used as tie-points for the combined BA (bundle adjustment) optimization. The perspective distortions and scale differences are the main issues that cause the failure of feature matching between aerial and ground images. According to the used strategy for alleviating these issues, existing methods can be divided into two categories, i.e., image rectification-based methods \citep{li2023fusion,liu2023tie,song2019oblique,wu2018integration} and view rendering-based methods \citep{gao2018accurate,gao2018ancient,zhu2020leveraging}. The former uses either rough POS data \citep{hu2015reliable,jiang2017board} or exploits common planes, such as building facades \citep{liu2023tie,wu2018integration}, for image global rectification. These methods can obviously decrease appearance differences and increase feature repeatability. However, their performance heavily relies on auxiliary data and scene structures. Thus, view rendering-based methods are further exploited, which mainly aim to render intermediate views to bridge the gap between aerial and ground images. Compared with image rectification-based methods, view rendering-based methods do not depend on specific scene structures. However, view rendering-based methods are degenerated by the high computational costs required to build dense and textured models and large distortions in rendered images.

In contrast to the above-mentioned hand-crafted methods, deep learning-based methods have also gained extensive attention in feature matching \citep{xu2024local}. Existing methods on one hand use the representation learning ability of CNN (convolutional neural network) for feature detection and description, which can be observed from the earlier networks for image patch description \citep{luo2019contextdesc,luo2018geodesc,mishchuk2017working,tian2017l2} to recent end-to-end networks for joint feature detection and description \citep{detone2018superpoint,dusmanu2019d2,luo2020aslfeat}. On the other hand, to enhance the discriminability of descriptors, GNN (graph neural network) has also been used in recent studies for feature matching, which obviously promotes the development of feature matching in challenging conditions, such as high texture-less regions and low illumination scenes \citep{chen2022aspanformer,edstedt2024roma,sarlin2020superglue,sun2021loftr,wang2022matchformer}. However, large perspective distortions are still the main challenge for feature matching of aerial and ground images. In recent years, deep learning has also promoted the development of novel view synthesis, such as the cutting-edge technique NeRF (Neural Radiance Field) \citep{mildenhall2021nerf} and 3DGS (3D Gaussian Splatting) \citep{kerbl20233d}. Compared with classical view rendering-based methods, these techniques have high rendering efficiency and synthesis quality because they can render views from sparse models without the time-consuming step for building dense models.

Inspired by these recent studies, this study proposes a reliable feature matching solution for aerial and ground images. Similar to view rendering-based methods, the core idea is to render intermediate views to alleviate perspective distortions and scale differences and build transmit matches between aerial and ground images. The main contributions of this study are summarized as follows: (1) we propose a view rendering algorithm via 3D Gaussian Splatting, which uses POS data for the coarse alignment of sparse models and calculation of intermediate view position and orientation; (2) we design a feature matching algorithm between aerial and ground images, which combines a deep learning-based feature matching network and transits aerial-ground feature matches through the rendered intermediate views; (3) we further verify the performance of the proposed solution by using real aerial and ground datasets in terms of feature matching and SfM (Structure from Motion) based 3D reconstruction.

This paper is organized as follows. Section 2 presents the related work in the literature. Section 3 gives the details of the proposed methodology for aerial and ground image matching, which is followed by experimental tests in terms of feature matching and 3D Gaussian Splatting scene rendering in Section 4. Finally, Section 5 presents conclusions and future studies.

\section{Related work}
\label{sec:2}
This study aims at feature matching between aerial and ground images. In the literature, existing methods can be grouped into three categories, i.e., image rectification-based methods, view rendering-based methods, and deep learning-based methods. Therefore, the following literature review will be conducted from these three aspects.

\subsection{Image rectification-based methods}
\label{sec:2.1}
The main difficulty in the feature matching of aerial and ground images is the perspective distortions and scale differences. Image rectification is the most direct strategy to decrease their geometric inconsistency. According to the used rectification strategy, existing methods can be divided into two groups, i.e., image horizontal rectification and image vertical rectification. The first one uses the POS data to rectify images onto geographic surfaces. By using onboard POS and ground elevation data, \citet{hu2015reliable} proposed rectifying aerial oblique images onto the same elevation plane, which decreases image deformations. In the work of \citet{jiang2017board}, a comparative evaluation has been conducted for UAV oblique images, which compares feature matching performance with and without image rectification. The tests demonstrate its validation in feature matching. In contrast to oblique image matching, the rectification plane is usually vertical considering the observation direction of aerial and ground images.

The second one is image vertical rectification. Instead of using an elevation plane, these methods exploit vertical primitives as projection planes. In urban environments, building facades are the most obvious plane primitives that have been exploited frequently in studies. Based on this idea, \citet{wu2018integration} proposed an approach for feature matching of aerial oblique imagery and terrestrial imagery. To address perspective distortion and scale variation, base planes are first selected from urban building facades that can be observed by both aerial and ground images. Image rectification is then executed to make building facades geometrically consistent in both images. The proposed feature matching method has been applied for high-quality 3D modeling of urban scenes. In their subsequent study \citep{li2023fusion}, the proposed solution has been further adopted for the integration of aerial, MMS, and backpack images and 3 mapping in urban areas. Instead of the above-mentioned image global rectification, \citet{liu2023tie} proposed an image patch rectification algorithm that rectifies image patches in one local normalized plane \citep{wu20083d}, which has been verified in feature matching of aerial and ground images. Compared with image global rectification, more time costs are consumed in generating and rectifying image patches. Generally, image rectification-based methods on the one hand depend on onboard POS data for geometric projection and coarse alignment; on the other hand, they rely on specific structures to support primitive plane selection.

\subsection{View rendering-based methods}
\label{sec:2.2}
View rendering-based methods aim to render intermediate views that are used to bridge the gap between aerial and ground images, which can reduce their perspective distortions and scale differences. Compared with image rectification-based methods, view rendering-based methods have two major advantages. On the one hand, they do not rely on specific structures to select primitive planes, such as building facades; on the other hand, they can render new images observing from any viewpoints. View rendering-based methods have gained attention in recent studies. As the earlier work, \citet{shan2014accurate} proposed a viewpoint-dependent matching method that uses depth maps and camera poses to warp ground images onto aerial images, and experiments for model geo-localization have been conducted to verify the feature matching solution. Based on the same idea, \citet{gao2018accurate} designed an accurate algorithm to align aerial and ground 3D models, whose core idea is to render aerial images from ground models at properly selected viewpoints. Considering outliers and noises in ground models, a view rendering algorithm has been designed to avoid artifacts and increase render quality.

In contrast to ground-to-aerial view rendering, \citet{zhu2020leveraging} proposed an aerial-to-ground view rendering algorithm that renders ground images from aerial models at ground views by using dense reconstructed mesh models and conducted feature matching between rendered images and ground images. With the aid of depths and normal vectors of rendered ground images, initial matches are then transmitted to aerial images. The proposed solution has been evaluated by using five open-sourced aerial-ground datasets. Due to fact that dense matching and texture mapping are extremely high time-consuming, the authors further designed a new view rendering strategy from sparse mesh models \citep{gao2018ancient}. Co-visible meshes were first extracted from ground mesh models, which were used to render aerial images by using triangle-induced homograph between aerial and ground images. The proposed solution has been verified by 3D modeling of ancient Chinese architectures. However, artifacts inevitably exist in the rendered images due to the usage of sparse model. In conclusion, although view rendering-based methods can bridge the gap between aerial and ground images, high time consumptions and low rendering quality are their major drawbacks.

\subsection{Deep learning-based methods}
\label{sec:2.3}
Due to the high representation learning ability, deep learning-based methods have gained rapid development in feature matching. Pioneer methods use CNN to extract high-level feature maps and implement feature detection and description. Existing networks can be categorized according to the steps of image matching, i.e., feature detection, feature description, and feature matching. The earlier networks aim at generating feature descriptors with high discriminative ability, such as L2-Net \citep{tian2017l2}, GeoDesc \citep{luo2018geodesc}, and ContextDesc \citep{luo2019contextdesc}, which feed into image patches and generate feature descriptors for the subsequent feature matching. Considering feature matching in difficult conditions, e.g., low-texture and low-illumination environments, joint feature detection and description networks have been proposed, which use image pairs as inputs and generate feature points and descriptors. The well-known networks include D2-Net \citep{dusmanu2019d2}, ASLFeat \citep{luo2020aslfeat}, and SuperPoint \citep{detone2018superpoint}. Joint feature detection and description networks use feature maps to detect keypoints and create descriptors. These methods have been extensively utilized for oblique image feature matching \citep{ji2023evaluation,jiang2021learned}.

Except for feature detection and description networks, GNN and Transformer-based attention mechanism have also been exploited in recent feature matching networks. SuperGlue \citep{sarlin2020superglue} is the most well-known network that adopts attentional GNN to enhance feature keypoints and descriptors into high-dimension vectors. Compared with L2 distance-based feature matching strategy, SuperGlue analogs human to find correspondences between image pairs and identify false matches. This work has greatly inspired following research. To cope with feature matching in non-texture regions, recent years have witnessed the arising of detector-free feature matching networks, e.g., LoFTR \citep{sun2021loftr}, ASpanFormer \citep{chen2022aspanformer}, and MatchFormer \citep{wang2022matchformer}. The core idea of these networks is to iteratively use self and cross attention layers in Transformer to obtain enhanced feature maps. For feature matching of aerial and ground images, detector-free networks have gain lots of attention in the field of photogrammetry \citep{li2023fusion,liang2023robust,zhu2024vdft}.

\section{Methodology}
\label{sec:3}
\begin{figure}[htbp]
    \centering
    \includegraphics[center]{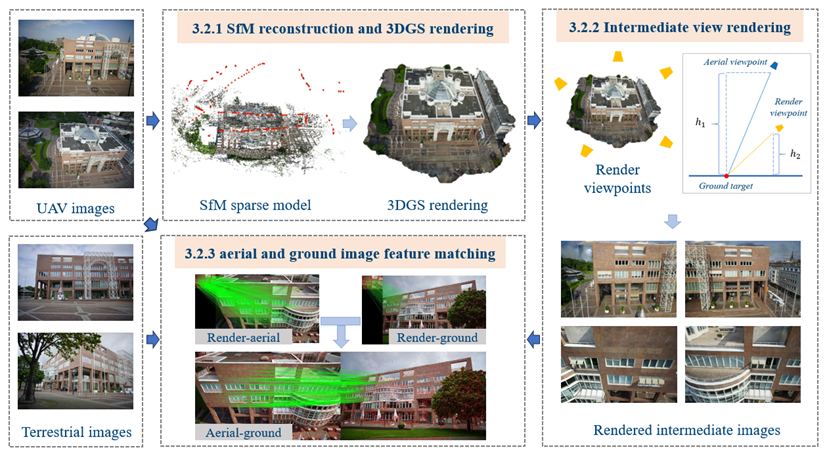}
    \caption{The workflow of the proposed algorithm.}
    \label{fig:1}
\end{figure}

Figure \ref{fig:1} shows the overall workflow of the proposed feature matching algorithm for aerial and ground images. UAV images and their rough POS data are the only inputs, and three major steps are conducted to achieve feature matching. First, a sparse model is constructed from aerial images through a well-known SfM pipeline, and the corresponding scene is rendered by using the 3DGS algorithm. Second, based on the SfM sparse model, a set of render viewpoints can be calculated according to aerial viewpoints and the pre-defined configurations, which are then used to guide the rendering of intermediate views from the 3DGS model. Third, by using the rendered images, feature matching is conducted among aerial, ground, and rendered images. That is, a set of feature matches can be obtained between rendered and aerial images as well as between rendered and ground images because of their relatively small perspective deformation. Finally, feature matches between aerial and ground images can be induced from the above results by using rendered images as the transistor.

\subsection{The principle of 3D Gaussian Splatting}
\label{sec:3.1}
The core idea of the proposed feature matching algorithm is using 3D Gaussian Splatting for photorealistic scene rendering and intermediate view generation. Compared with classical scene rendering solutions such as MVS (Multi-View Stereo) based dense matching and texture mapping, 3D Gaussian Splatting has the advantage of low-memory consumption, high-speed and high-quality rendering. Thus, the principle of 3D Gaussian Splatting is first presented.

3D Gaussian Splatting is an explicit radiance field representation technique. It represents a scene by using a large set of 3D anisotropic balls, i.e., 3D Gaussians that capture the desirable properties of continuous volumetric radiance fields for scene optimization \citep{kerbl20233d}. Each 3D Gaussian $G(x)$, as presented by Equation \ref{equ:1}, is characterized by its mean vector $\mu$, covariance matrix $\sum$, opacity $\alpha$, and color $c$ that is calculated from spherical harmonics as view-dependent appearance. The covariance matrix $\sum$ can be decomposed into the rotation matrix $R \in \mathbb{R}^{3 \times 3}$ and the scale matrix $S \in \mathbb{R}^{3 \times 1}$, i.e., $\sum=R S S^T R^T$.

\begin{equation}
    G(x)=e^{-\frac{1}{2}(x-\mu)^T \Sigma^{-1}(x-\mu)} 
    \label{equ:1}
\end{equation}

When blended, 3D Gaussians create a full model that can be rendered from any viewpoint, which can support novel view rendering. The input of 3D Gaussian Splatting is a sparse model with point clouds and oriented images generated from SfM. Each 3D point in the point cloud is converted into a 3D Gaussian, with the mean and covariance of the Gaussian corresponding to the point's position and uncertainty. This results in a set of Gaussians that are then optimized during training to minimize the difference between the rasterized images and the real images. For view rendering, these Gaussians are first projected onto the 2D image plane, and their color is calculated by using spherical harmonics. After sorted according to depth, the color of a pixel $x$ is calculated by the $\alpha$-blending of opacity and color of Gaussians that cover the pixel by depth order, as presented by Equation \ref{equ:2}, where $c_i$ is the learned color; $\alpha_i^{\prime}$ is calculated by the learned opacity $\alpha_i$ and the Gaussian, i.e., $\alpha_i^{\prime}=\alpha_i G(X)$.

\begin{equation}
    C(x)=\sum_{i \in N} \alpha_i^{\prime} c_i \prod_{j=1}^{i-1}\left(1-\alpha_j^{\prime}\right)    
    \label{equ:2}
\end{equation}

3D Gaussian Splatting does not depend on dense sampling, which is very computationally expensive and memory-intensive, as in NeRF \citep{mildenhall2021nerf}. Instead, 3D Gaussian Splatting uses a tile-based rasterizer for efficient volume rendering, which can quickly sort and handle occlusions while limiting the number of Gaussians that receive gradients. It allows for high-quality novel view synthesis with only a few input views and has been shown to achieve real-time rendering speeds with competitive training times and visual quality. Therefore, 3D Gaussian Splatting is adopted in this study for intermediate view rendering to bridge the gap between aerial and ground images.

\subsection{Intermediate view rendering for feature matching}
\label{sec:3.2}
According to the procedure of 3D Gaussian Splatting, this section presents the proposed intermediate view rendering algorithm for feature matching of aerial and ground images. The overall workflow consists of three major steps, including aerial image SfM reconstruction, view rendering between aerial-ground images, and aerial-ground image feature matching.

\subsubsection{Aerial image SfM reconstruction}
\label{sec:3.2.1}
3D Gaussian Splatting uses a sparse model as its input, which includes sparse point clouds and oriented images. The former is used to initialize Gaussians; the latter is utilized to calculate training loss and refine rendering scenes. In the literature, incremental SfM (ISfM) has become the golden standard technique for sparse model generation, especially for UAV images \citep{jiang2020efficient}, because of its high robustness to feature matching outliers and low dependency on initial values of unknown parameters. Thus, as in 3D Gaussian Splatting \citep{kerbl20233d}, the widely used open-source library ColMap \citep{schonberger2016structure} with its default configurations has been used for generating sparse models.

\begin{figure}[htbp]
\centering
    \subfloat[]{\includegraphics[width = 0.6\textwidth]{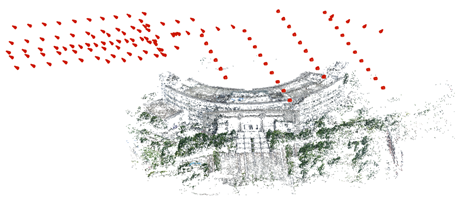}}
    \hfill
    \subfloat[]{\includegraphics[width = 0.4\textwidth]{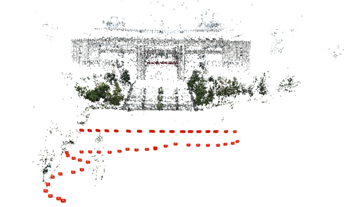}}
    \caption{The generated sparse model by using (a) aerial images and (b) ground images. Image planes are rendered as red rectangles, and 3D points are rendered with image colors.}
    \label{fig:2}
\end{figure}

For the proposed aerial-ground feature matching workflow, there are two options to select images for generating sparse models, i.e., aerial images and ground images. Generally, aerial images are captured from high altitudes with complete scene coverage, as shown in Figure \ref{fig:2}(a); on the contrary, ground images are collected with near distances, which can record more scene detail, as presented in Figure \ref{fig:2}(b). To ensure that the subsequently rendered images have large and valid overlap regions with existing images, aerial images are selected to conduct the ISfM reconstruction and generate the sparse model.

\subsubsection{View rendering between aerial-ground images}
\label{sec:3.2.2}
Photorealistic scene rendering can be achieved based on the reconstructed sparse model. In the literature, classical SfM-MVS based workflows sequentially execute dense matching for point cloud generation, point meshing for Delaunay triangulation construction, and texture mapping for real scene rendering. These solutions have the disadvantages of time-consuming and low quality due to the long procedure and high computational burden. Thus, this study uses 3D Gaussian Splatting for scene rendering.

\begin{figure}[htbp]
    \centering
    \subfloat[]{\includegraphics[width = 0.48\textwidth]{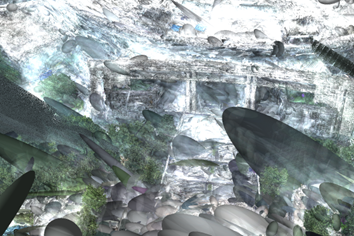}}
    \hfill
    \subfloat[]{\includegraphics[width = 0.48\textwidth]{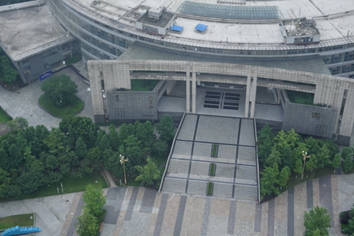}}
    \caption{The visualization of (a) 3D Gaussians in the rendered scene and (b) novel view image rendering based on 3D Gaussian Splatting.}
    \label{fig:3}
\end{figure}

The purpose of scene rendering is to rapidly generate high-quality intermediate images. Considering the high resolution of aerial images, the training time and required GPU memory are substantial. However, 3D Gaussian Splatting can still produce high-quality 3D models even under low-resolution conditions. In this study, the original images are first down-sampled with the scaling width and height proportionally to a width of less than 1,500 pixels. Taking as input the camera poses and 3D point clouds from the SfM sparse model, 3D Gaussian Splatting is used for iterative training to render the entire scene, as shown in Figure \ref{fig:3}.

\begin{figure}[htbp]
    \centering
    \includegraphics{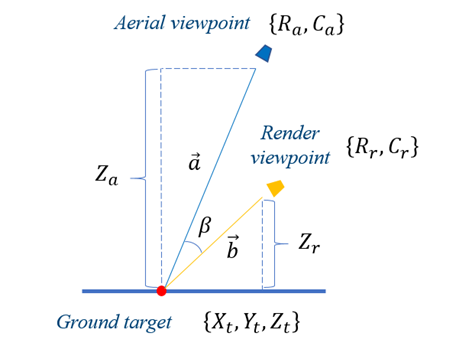}
    \caption{The illustration of render viewpoint determination based on the oriented camera poses of aerial images from SfM reconstruction.}
    \label{fig:4}
\end{figure}

Render viewpoint determination is critical for aerial-ground intermediate view rendering as it mainly affects the perspective deformations between rendered images and aerial-ground images. By using the SfM reconstruction, this study designs an algorithm to determine render viewpoints, as illustrated in Figure \ref{fig:4}. Assuming that the camera pose of an aerial viewpoint is represented as $\left\{R_a, C_a\right\}$, in which $R_a$ and $C_a=\left\{X_a, Y_a, Z_a\right\}$ indicate the rotation matrix and projection center, respectively. Based on the imaging geometry, the coordinate $\left\{X_t, Y_t, Z_t\right\}$ of the ground target observed through the principle point of the aerial viewpoint is calculated via ray intersection. The core idea of the proposed algorithm is that the rendered images must have large overlap regions with aerial and ground images and have as less as possible perspective deformations. Thus, the render viewpoint is determined by moving the position of the aerial viewpoint along the vertical direction and rotating the camera around $X$ axis until the ground target is observed again via the principle point of the rendered viewpoint. According to above definition, the camera pose $\left\{R_r, C_r\right\}$ of the render view points is calculated via Equations \ref{equ:3} and \ref{equ:4} , in which $\beta$ is the intersection angle of vectors $\vec{a}$ and $\vec{b}$, i.e., $\cos (\beta)=\vec{a} * \vec{b} /\|\vec{a}\|\|\vec{b}\|$. By using these render viewpoints, intermediate images can be generated from the 3D GS model.

\begin{equation}
    \left\{\begin{array}{c}
    X_r=X_a \\
    Y_r=Y_a \\
    Z_r=\frac{1}{2} Z_a
    \end{array}\right. 
    \label{equ:3}
\end{equation}

\begin{equation}
    R_r=R_a * R_X(-\beta)
    \label{equ:4}
\end{equation}

\subsubsection{Aerial-ground image feature matching}
\label{sec:3.2.3}
Aerial and ground images have significant differences in viewing angles and low overlaps, which leads to a large variance in feature descriptors and low repeatability of corresponding points. It in turn reduces the reliability and accuracy of direct feature matching. The principle of the proposed algorithm in this paper is to introduce intermediate images between aerial and ground images. The intermediate images reduce perspective differences and increase overlap regions between aerial-ground images, which decreases the difficulty of direct aerial-ground feature matching and serves as a bridge to assist in feature matching.

By using the rendered intermediate images, feature matching can then be executed using a well-known algorithm. In this study, we have tested using SIFT with default configurations. It is shown that the default SIFT method resulted in very few initial matches among the three sets of images, which could not meet the requirements of subsequent experiments. Therefore, recent deep learning-based methods, i.e., SuperPoint \& SuperGlue \citep{detone2018superpoint,sarlin2020superglue}, have been selected for feature extraction and matching. First, SuperPoint is used for local feature extraction from all images to obtain all local features. Second, SuperGlue is adopted to execute feature matching between intermediate images and aerial-ground images. Finally, after outlier removal based on fundamental matrix estimation, final matches can be obtained for aerial-intermediate and ground-intermediate image pairs. The results reveal that the feature matching solution can achieve a larger match number and higher precision.

\begin{figure}[htbp]
    \centering
    \includegraphics[width = 0.8\textwidth]{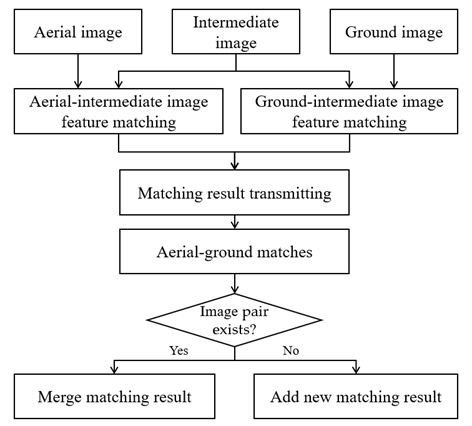}
    \caption{The procedure of aerial-ground feature matching and merging.}
    \label{fig:5}
\end{figure}

Feature matches between aerial-ground images should be extracted after obtaining feature matching results of aerial-intermediate images and ground-intermediate images. The process of merging feature matching results between intermediate images and aerial-ground images is shown in Figure \ref{fig:5}. The main procedure consists of the steps: (1) for one intermediate image, reading its matching pairs with aerial and ground images, i.e., one pair of aerial-intermediate image matches and one pair of ground-intermediate image matches; (2) extracting match data from these two image pairs, and merging them to generate aerial-ground image matching pairs; (3) checking if the new aerial-ground image pair whether or not exists; if it does, merging the new and old feature matching results and ensure their uniqueness; if not, directly establishing a new image pair; (4) repeating the above steps for all intermediate images. Especially, step (2) involves reading match data corresponding to aerial-intermediate image pairs and ground-intermediate image pairs and storing them in Array 1 and Array 2, respectively. Then, finding the points in the intermediate image that exist in both Array 1 and Array 2, and storing them as common points, which are the matching points of aerial and ground images. According to these processing steps, final matches between aerial-ground images can be obtained.

\subsection{Algorithm implementation}
\label{sec:3.3}
According to the proposed image matching workflow, this study has implemented the algorithm by using open-source and self-developed packages. Especially, the widely used ISfM toolkit ColMap with the default configurations has been used for generating sparse models, in which 3D points and oriented images are used for 3D Gaussian Splatting based scene rendering. For 3DGS-based scene rendering, we set the total number of training iterations as 30k; we also update densification every 100 times and reset opacity every 3,000 times. In addition, we set opacity, scale, and rotation to 0.05, 0.005, and 0.001, respectively. For aerial-ground image feature matching, we set the input image size to 1440 pixels in SuperPoint \& SuperGlue and set the maximum keypoint number to 8192, which is similar to SIFTGPU in ColMap.

\begin{table}[htbp]
    \centering
    \caption{The source list of used software packages.}
    \begin{tabular}{cl}
    \hline
    \textbf{Software package} & \multicolumn{1}{c}{\textbf{Website source}}                      \\ \hline
    ColMap                    & https://github.com/colmap/colmap                        \\
    3DGS                      & https://github.com/graphdeco-inria/gaussian-splatting   \\
    SuperPoint                & https://github.com/rpautrat/SuperPoint                  \\
    SuperGlue                 & https://github.com/magicleap/SuperGluePretrainedNetwork \\ \hline
    \end{tabular}
    \label{tab:1}
\end{table}

\section{Experimental results}
\label{sec:4}
In the experiments, three aerial-ground datasets are used to evaluate the performance of the proposed algorithm. First, we compare the efficiency and quality of intermediate views that are rendered by classical SfM-MVS and the 3DGS-based solutions. Second, feature matching is then conducted to evaluate the performance of the aerial-ground image matching algorithm, which is compared with other two feature matching methods. Third, the matching results are utilized in ISfM reconstruction to validate its practical applications. In this study, the tests are conducted on a Windows platform that is equipped with 64 GB memory, a 3.0 GHz Intel Core i9-13900K CPU and a 24 GB NVIDIA GeForce RTX 4090 graphic card.

\subsection{Test sites and datasets}
\label{sec:4.1}
\begin{figure}[htbp]
\centering
    \subfloat[CENTER, pairs 1-4]{\includegraphics[width = 0.32\textwidth]{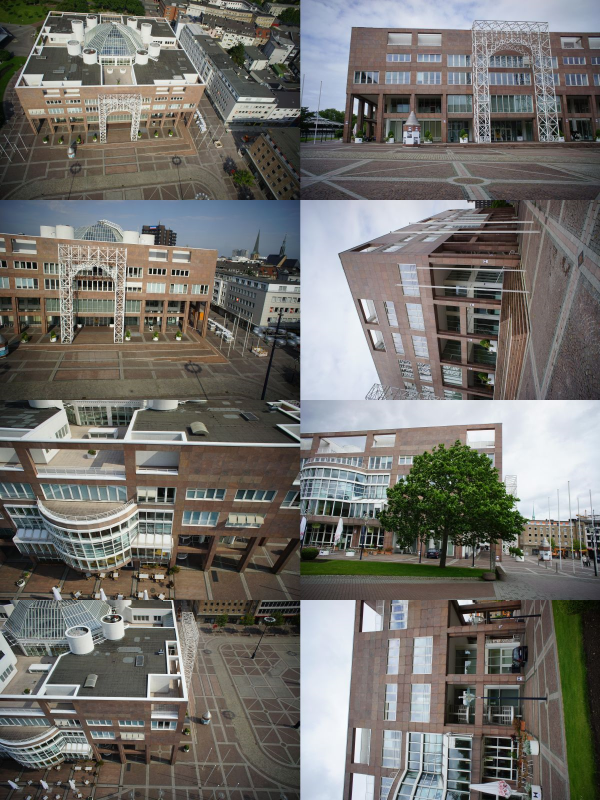}}
    \hfill
    \subfloat[ZECHE, pairs 5-8]{\includegraphics[width = 0.32\textwidth]{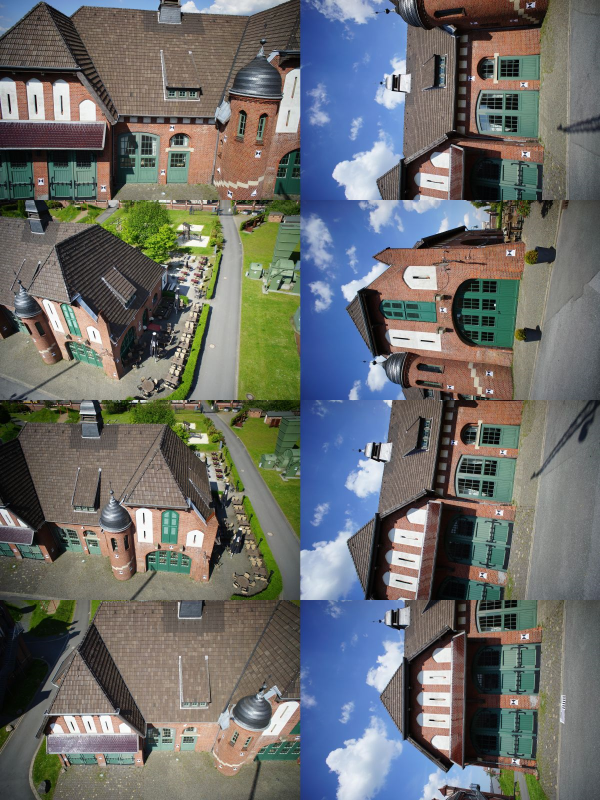}}
    \hfill
    \subfloat[SWJTU-LIB, pairs 9-12]{\includegraphics[width = 0.32\textwidth]{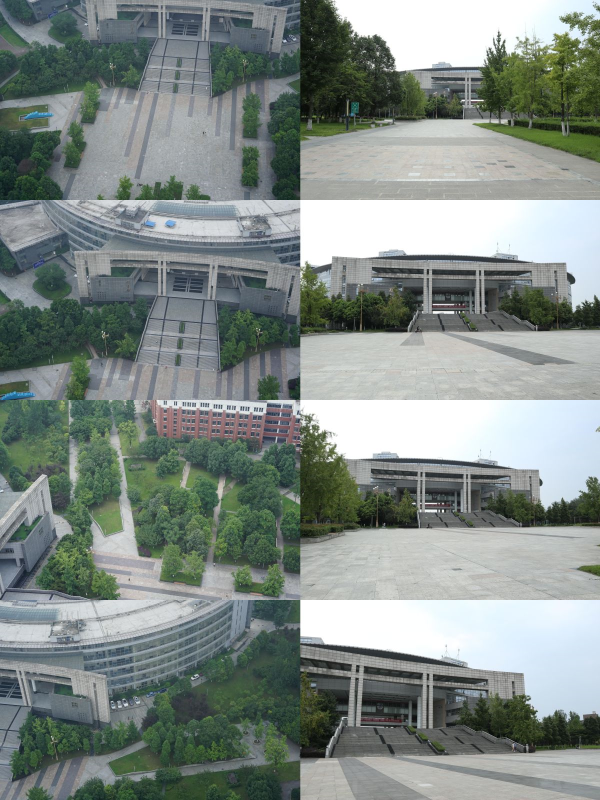}}
    \caption{The selected 12 image pairs from the three test datasets. The left and right columns severally show aerial and ground images from their corresponding datasets.}
    \label{fig:6}
\end{figure}

For performance evaluation, three widely used aerial-ground datasets have been used in this study. The details of these datasets are listed in Table \ref{tab:2}. The first and second datasets, i.e., CENTER and ZECHE, are collected and prepared as the “ISPRS benchmark for multi-platform photogrammetry”, which serves as a multi-sensor benchmark to evaluate image orientation and dense matching algorithms \citep{nex2015isprs}. Both aerial images and ground images in these two datasets are recorded by using Sony NEX-7 cameras with dimensions of 6,000 by 4,000 pixels. For the dataset Center, there are 146 and 204 images collected from aerial and ground sensors with the GSD (ground sampling distance) of 1.10 cm and 0.53 cm, respectively; for the dataset ZECHE, there are 172 and 147 aerial and ground images, whose GSD respectively reaches to 0.56 cm and 0.28 cm due to lower flight altitudes. The third dataset is collected from a campus library building \citep{zhu2020leveraging} by using Sony ICLE 510 and Cannon EOS M6 cameras for aerial and ground image collection, respectively. There are 123 and 78 images recorded from the aerial and ground sensors, whose GSD is 1.69 cm and 1.06 cm, respectively. For these three datasets, a total number of 12 image pairs have been selected to evaluate the performance of feature matching, as shown in Figure \ref{fig:6}.

\begin{table}[htbp]
    \centering
    \caption{Detailed information of the three datasets.}
    \begin{tabular}{lllllll}
    \hline
    \multirow{2}{*}{\textbf{Item Name}} & \multicolumn{2}{c}{\textbf{CENTER}}                                                                             & \multicolumn{2}{c}{\textbf{ZECHE}}                                                                                & \multicolumn{2}{c}{\textbf{SWJTU-LIB}}                                                                               \\ \cline{2-7} 
                                        & \textbf{Aerial}                                         & \textbf{Ground}                                       & \textbf{Aerial}                                         & \textbf{Ground}                                         & \textbf{Aerial}                                          & \textbf{Ground}                                           \\ \hline
    Camera mode                         & \begin{tabular}[c]{@{}l@{}}Sony   \\ NEX-7\end{tabular} & \begin{tabular}[c]{@{}l@{}}Sony \\ NEX-7\end{tabular} & \begin{tabular}[c]{@{}l@{}}Sony   \\ NEX-7\end{tabular} & \begin{tabular}[c]{@{}l@{}}Sony   \\ NEX-7\end{tabular} & \begin{tabular}[c]{@{}l@{}}Sony \\ ICLE 510\end{tabular} & \begin{tabular}[c]{@{}l@{}}Cannon  \\ EOS M6\end{tabular} \\
    GSD (cm)                            & 1.10                                                    & 0.53                                                  & 0.56                                                    & 0.28                                                    & 1.69                                                     & 1.06                                                      \\
    Number of   images                  & 146                                                     & 203                                                   & 172                                                     & 147                                                     & 123                                                      & 78                                                        \\
    With POS   data                     & Yes                                                     & Yes                                                   & Yes                                                     & Yes                                                     & Yes                                                      & Yes                                                       \\
    Image   size (pixel)                & \multicolumn{2}{l}{6,000×4,000}                                                                                 & \multicolumn{2}{l}{6,000×4,000}                                                                                   & \multicolumn{2}{l}{6,000×4,000}                                                                                      \\ \hline
    \end{tabular}
    \label{tab:2}
\end{table}

\subsection{Evaluation and metrics}
\label{sec:4.2}
The performance of the proposed algorithm would be evaluated in feature matching and ISfM reconstruction. The former uses 12 image pairs that are selected from the three datasets to evaluate and compare its performance in feature matching; the latter feeds as input feature matching results and conducts ISfM reconstruction for aerial and ground images, which can then be used to generate complete 3D models. For the evaluation of feature matching, three metrics are used, including NCM (number of correct matches), NIM (number of initial matches), and NMP (number of match pairs). NCM indicates the number of true matches that are retained after outlier removal; NIM indicates the number of matches that pass through nearest neighbor searching and ratio test; NMP is the number of match pairs that have at least 15 true matches. For the evaluation of ISfM reconstruction, two metrics are also selected, i.e., NRI (number of registered images) and N3P (number of 3D points). NRI is the number of images that are successfully registered in image orientation; N3P represents the number of resumed 3D points. The details of conducted evaluation and used metrics are shown in Table \ref{tab:3}.

\begin{table}[htbp]
    \centering
    \caption{The details of conducted evaluation and used metrics.}
    \begin{tabular}{lcl}
    \hline
    \textbf{Evaluation}       & \textbf{Metric} & \textbf{Description } \\    \hline
    \multirow{3}{*}{Feature   matching}  & NCM    & The number of true matches.        \\
                                         & NIM    & The number   of matches.           \\
                                         & NMP    & The number   of match pairs.       \\ \hline
    \multirow{2}{*}{ISfM reconstruction} & NRI    & The number   of registered images. \\
                                         & N3P    & The number   of resumed 3D points. \\ \hline
    \end{tabular}
    \label{tab:3}
\end{table}

\subsection{Motivation for Rendering Intermediate Views with 3DGS}
\label{sec:4.3pre}
This study opts to utilize 3DGS for rendering intermediate views rather than directly employing 3DGS representations for pose optimization. This decision is grounded in an in-depth analysis of existing pose-free 3DGS methods. Specifically, there are significant viewpoint discrepancies between aerial and ground images, encompassing aspects such as altitude, lighting, and texture distribution. These differences complicate direct pose optimization on implicit representations. In contrast, rendering intermediate views simplifies the problem to a scenario more akin to 2D image matching, thereby handling these discrepancies more effectively.

Despite the existence of some excellent pose-free 3DGS methods, such as CF3dgs \citep{fu2024colmap}, NoPoSplat \citep{ye2024no}, PF3plat \citep{hong2024pf3plat}, etc., they exhibit limitations when dealing with open area tasks. These methods perform well in smaller scenes, for instance, capturing hundreds of images of a single object can yield satisfactory results. However, in the large open area scenes involved in this paper, due to the low overlap between images, these methods struggle to complete effective reconstructions. Moreover, these methods are computationally demanding and have primarily been tested on low-resolution images; for example, NoPoSplat and PF3plat mainly use image datasets of 256×256, and even CF3dgs only handles images up to 960×540.

\begin{figure}[htbp]
    \centering
    \includegraphics[width = \textwidth]{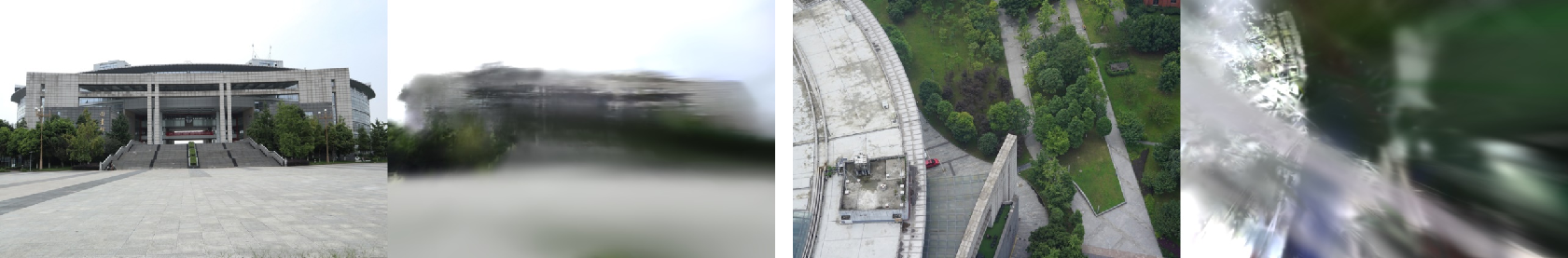}
    \caption{The illustration of CF3dgs scene rendering.}
    \label{fig:add0}
\end{figure}

To further verify the performance of pose-free 3DGS methods in processing open area scenes, we tested the SWJTU-LIB dataset. The dataset's original resolution is 6000×4000, which we downscaled by a factor of eight to 750×500 for testing with the CF3dgs method. The test results indicate that the average Peak Signal-to-Noise Ratio (PSNR) of the images rendered using the CF3dgs method is only 15. This value is significantly below the image quality standards typically considered acceptable, indicating a considerable discrepancy between the rendered images and the original images. Additionally, visually inspecting the rendered results in Figure \ref{fig:add0}, we can observe noticeable blurring and distortion in the images, particularly around edges and texture details. This further confirms the numerical results that CF3dgs may not achieve the desired rendering quality when handling open area scenes. Therefore, this study adopts the method of rendering intermediate views with 3DGS to overcome the aforementioned challenges.

\subsection{Analysis of the performance for image rendering}
\label{sec:4.3}
Image rendering is the critical step in the workflow of the proposed aerial-ground feature matching algorithm. In general, image rendering can be implemented by using the MVS-based pipeline that sequentially conducts image orientation, dense matching, point meshing, and texture mapping. Main issues arise from the high time cost of dense matching and low quality of scene rending. On the contrary, 3DGS exploits neural radiance field for scene representation, which can achieve efficient scene rendering with high-quality details. Thus, this section will compare the performance of these two image rendering solutions.

\begin{figure}[htbp]    
\centering
    MVS-based rendering \hspace{2.5cm} 3DGS-based rendering   
    \subfloat[CENTER]{\includegraphics{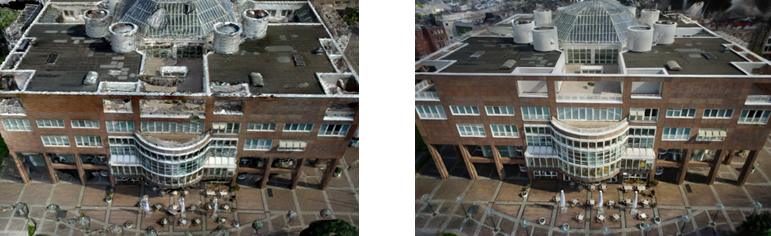}}
    \\
    \subfloat[ZECHE]{\includegraphics{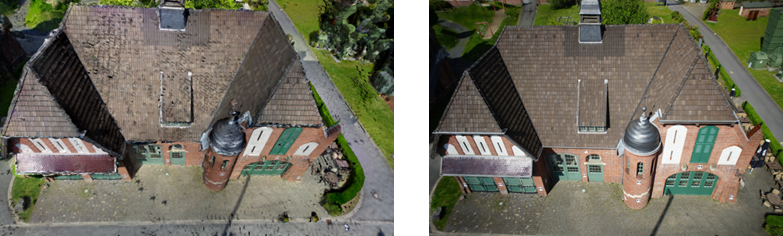}}
    \\
    \subfloat[SWJTU-LIB]{\includegraphics{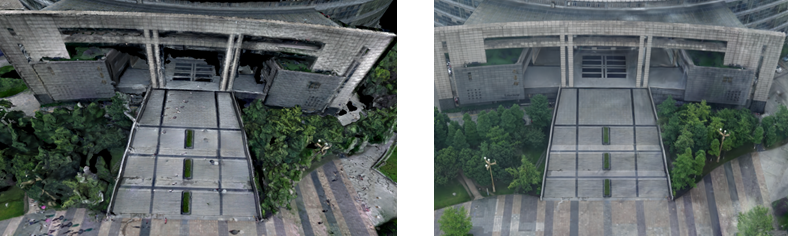}}
    \caption{The illustration of intermediate view rendering for the three datasets.}
    \label{fig:7}
\end{figure}

By using these three datasets, MVS and 3DGS are utilized for performance evaluation, and these two solutions take as input sparse models that are reconstructed using an ISfM pipeline. Noticeably, only aerial images are used in intermediate view rendering in this test. The metric efficiency is used for performance evaluation, which is the time cost of scene rendering. The results are listed in Table \ref{tab:4}. Wen can see that 3DGS consumes 20.0 min, 25.0 min, and 28.0 min in scene rendering for the three datasets, respectively. However, the time cost of MVS is about 2.75, 2.16, and 3.07 times of that in 3DGS. In addition, the quality of scene rendering of MVS is much lower than 3DGS, as illustrated in Figure \ref{fig:7}, in which the left column indicates the results of MVS-based rendering; the right column represents the results of 3DGS-based rendering. It is clearly shown that there are many erroneous regions observed from the results of MVS, e.g., building roofs with texture-less regions and repetitive patterns and building facades that have transparent glasses. These regions pose challenges on MVS-based scene rendering method. On the contrary, 3DGS can give photorealistic rendering results and can facilitate the subsequent feature matching of aerial and ground images via intermediate view rendering.

\begin{table}[htbp]
    \centering
    \caption{The efficiency of different rendering solutions (unit in minutes).}
    \begin{tabular}{crr}
    \hline
    \textbf{Dataset} & \textbf{MVS} & \textbf{3DGS} \\ \hline
    CENTER           & 55.0 & 20.0 \\
    ZECHE            & 54.0 & 25.0 \\
    SWJTU-LIB        & 86.0 & 28.0 \\ \hline
    \end{tabular}
    \label{tab:4}
\end{table}

\begin{figure}[htbp]    
\centering 
    \subfloat[MVS-based rendering]{\includegraphics[width = \textwidth]{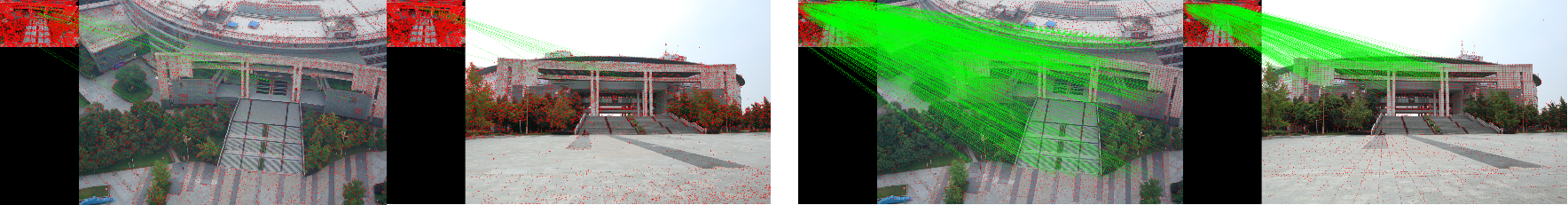}}
    \\
    \subfloat[3DGS-based rendering]{\includegraphics[width = \textwidth]{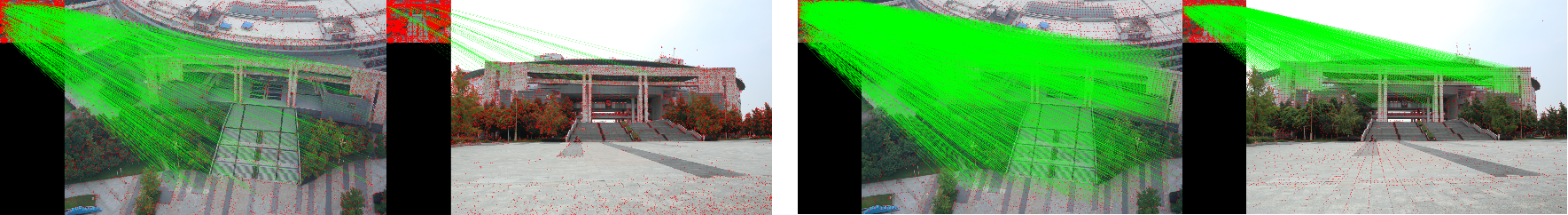}}
    \caption{The illustration of different rendering solutions in feature matching.}
    \label{fig:add1}
\end{figure}

To further evaluate the practical performance of the two image rendering solutions in feature matching, we selected rendered images from the SWJTU-LIB dataset and the image pair 9 for feature matching tests. We employed two feature matching methods: SIFTGPU and SuperPoint \& SuperGlue. Figure \ref{fig:add1} shows the visualization results of the two rendering methods, while Table \ref{tab:add1} lists the specific quantitative outcomes, including NCM (Number of Correct Matches) and NIM (Number of Initial Matches).

The visualization results in Figure \ref{fig:add1} clearly demonstrate the advantages of the 3DGS-based rendering method in feature matching. Specifically, the images rendered with 3DGS provide a greater number of correct matches (NCM) and initial matches (NIM) during the feature matching process, which is beneficial for enhancing the accuracy and efficiency of subsequent feature matching. In contrast, although the MVS-based rendering method can also perform feature matching, it falls short of the 3DGS method in both the quantity and quality of matching points. The quantitative results in Table \ref{tab:add1} further confirm this. These results indicate that the 3DGS rendering method has a significant advantage in feature matching, capable of providing richer and more accurate feature match information for subsequent 3D reconstruction. This not only verifies the superiority of 3DGS in rendering quality and feature matching performance but also provides strong support for its future application in complex scenarios.

\begin{table}[htbp]
    \centering
    \caption{Quantitative results on image pair 9. The values indicate NCM and NIM.}
    \begin{tabular}{lllll}
    \hline
    \multirow{2}{*}{\textbf{Image rendering}} & \multicolumn{2}{c}{\textbf{SIFTGPU}} & \multicolumn{2}{c}{\textbf{SuperPoint \& SuperGlue}} \\ \cline{2-5} 
                                              & \textbf{Aerial}   & \textbf{Ground}  & \textbf{Aerial}           & \textbf{Ground}          \\ \hline
    MVS-based                                 & 40/67             & 20/35            & 1213/1580                 & 490/603                  \\
    3DGS-based                                & 928/965           & 59/78            & 3159/3214                 & 984/1065                 \\ \hline
    \end{tabular}
    \label{tab:add1}
\end{table}

\subsection{Evaluation of aerial and ground feature matching}
\label{sec:4.4}
With the aid of intermediate views, feature matching can then be conducted for aerial and ground images, which alleviates viewpoint changes and transmits feature correspondences. By using 12 image pairs selected from three datasets as shown in Figure 6, this section analyzes the performance of the proposed feature matching algorithm in terms of NCM and NIM. For performance comparison, two widely used solutions, i.e., the classical algorithm SIFTGPU and the deep learning network SuperPoint \& SuperGlue, have also been evaluated.

\begin{figure}[htbp]
    \centering
    \subfloat[SIFTGPU (0/27)]{\includegraphics{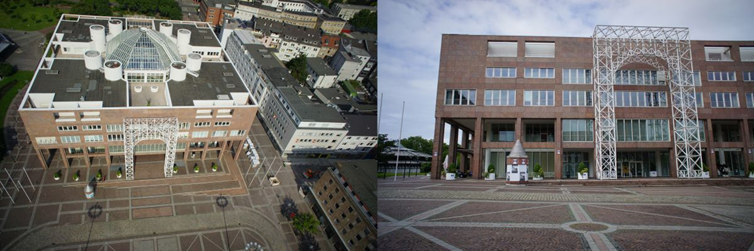}}
    \\
    \subfloat[SuperPoint \& SuperGlue (143/236)]{\includegraphics{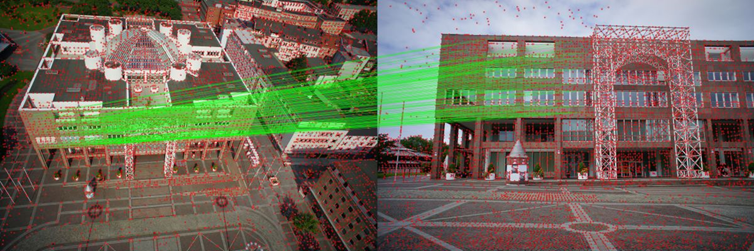}}
    \\
    \subfloat[Intermediate view aided feature matching]{\includegraphics{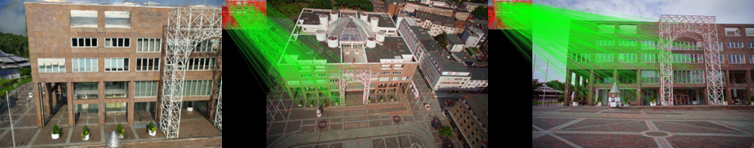}}
    \\
    \subfloat[Ours (845/2,074)]{\includegraphics{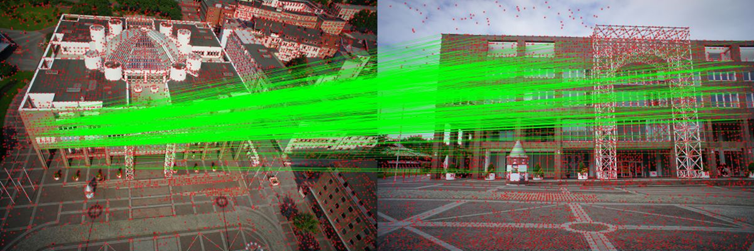}}
    \caption{The illustration of aerial-ground matching results (image pair 1) for dataset CENTER. The values in the bracket indicate the metric NCM and NIM, respectively.}
    \label{fig:8}
    \vspace{-16.60066pt}
\end{figure}

\begin{figure}[htbp]
    \centering
    \subfloat[SIFTGPU (0/20)]{\includegraphics{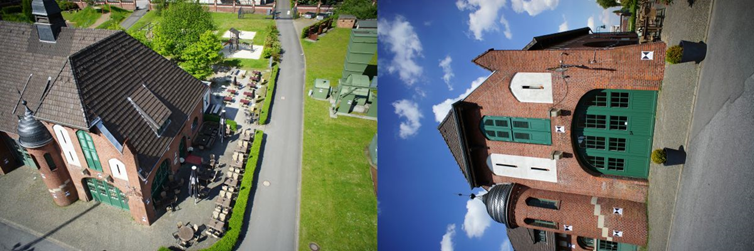}}
    \\
    \subfloat[SuperPoint \& SuperGlue (0/8)]{\includegraphics{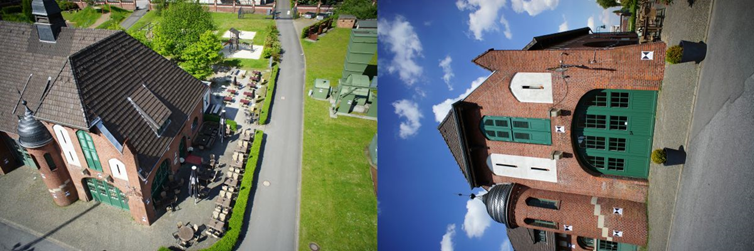}}
    \\
    \subfloat[Intermediate view aided feature matching]{\includegraphics{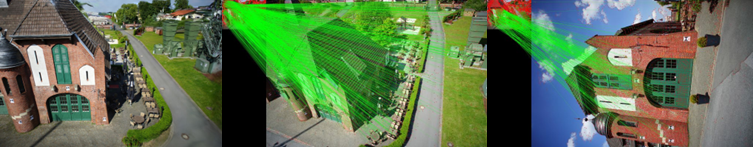}}
    \\
    \subfloat[Ours (108/924)]{\includegraphics{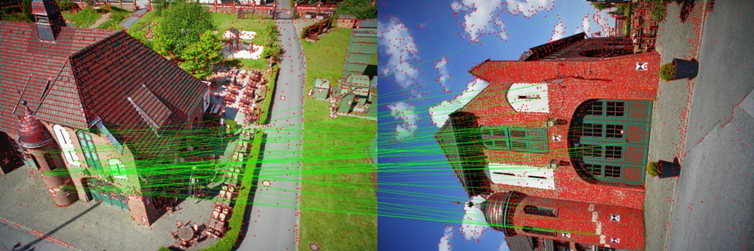}}
    \caption{The illustration of aerial-ground matching results (image pair 5) for dataset ZECHE. The values in the bracket indicate the metric NCM and NIM, respectively.}
    \label{fig:9}
    \vspace{-16.11887pt}
\end{figure}

Table \ref{tab:5} presents the quantitative results on the selected image pairs, in which there are 4 images selected from each dataset. It is clearly shown that because of large viewpoint changes, SIFTGPU failed in feature matching for many image pairs, e.g., 1, 2, and 4 in CENTER and 5, 6, and 7 in ZECHE, and 10 in SWJTU-LIB. The main reason is that there are very few initial matches that can be found from aerial and ground images with large appearance differences, as illustrated in Figure \ref{fig:8}(a) and Figure \ref{fig:9}(a). By using deep learning-based feature detection and description, SuperPoint \& SuperGlue obviously increases the performance of feature matching for aerial and ground images, especially for CENTER and SWJTU-LIB, when compared with SIFTGPU. However, for dataset ZECHE, feature matching still failed in image pairs 5, 6, and 7. The main reason is the extremely large viewpoint changes, including the perspective distortion and appearance rotation, as shown in Figure \ref{fig:9}(b).

\begin{table}[htbp]
    \centering
    \caption{Quantitative results on selected image pairs. The values indicate NCM and NIM.}
    \begin{tabular}{crrr}
    \hline
    \textbf{Image pair} & \textbf{SIFTGPU} & \textbf{SuperPoint \& SuperGlue} & \textbf{Ours} \\ \hline
    1                   & 0/27             & 143/236                          & 845/2,074     \\
    2                   & 0/0              & 105/144                          & 223/626       \\
    3                   & 24/45            & 240/438                          & 714/2,001     \\
    4                   & 0/35             & 0/0                              & 174/404       \\ \hline
    5                   & 0/20             & 0/8                              & 108/924       \\
    6                   & 0/0              & 0/0                              & 216/1,813     \\
    7                   & 0/0              & 0/0                              & 183/856       \\
    8                   & 79/109           & 145/3,054                        & 326/558       \\ \hline
    9                   & 146/173          & 820/1,074                        & 2,547/13,057  \\
    10                  & 0/16             & 57/101                           & 182/663       \\
    11                  & 42/68            & 400/483                          & 1,268/4,035   \\
    12                  & 35/50            & 225/333                          & 669/1,282     \\ \hline
    \end{tabular}
    \label{tab:5}
\end{table}

\begin{figure}[htbp]
    \centering
    \subfloat[SIFTGPU (146/173)]{\includegraphics{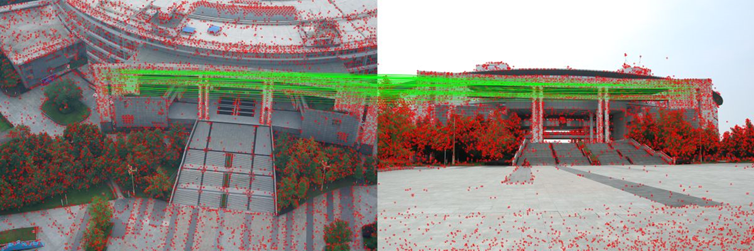}}
    \\
    \subfloat[SuperPoint \& SuperGlue (820/1,074)]{\includegraphics{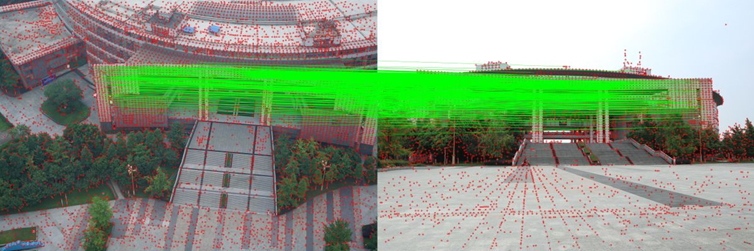}}
    \\
    \subfloat[Intermediate view aided feature matching]{\includegraphics{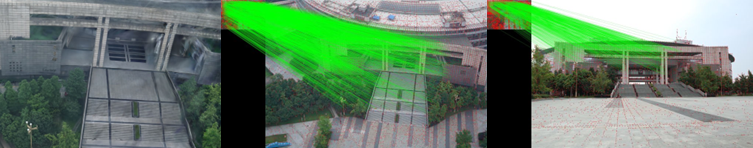}}
    \\
    \subfloat[Ours (2,547/13,057)]{\includegraphics{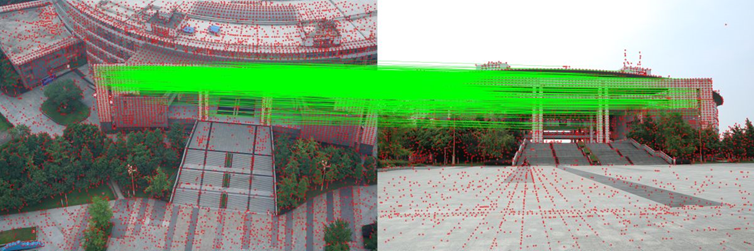}}
    \caption{Aerial-ground matching results (image pair 9) for dataset SWJTU-LIB. The values in the bracket indicate the metric NCM and NIM, respectively.}
    \label{fig:10}
    \vspace{-16.60066pt}
\end{figure}

On the contrary, for the proposed algorithm, reliable feature matching can be achieved for all image pairs. When compared with SuperPoint \& SuperGlue, the metrics NCM and NIM are obviously better, especially for dataset ZECHE. The main reason is the usage of intermediate views. For further analysis, Figure \ref{fig:8}(c), Figure \ref{fig:9}(c), and Figure \ref{fig:10}(c) show rendered images, feature matching results of render-aerial and render-ground image pairs, respectively, and the final matches are presented in Figure \ref{fig:8}(d), Figure \ref{fig:9}(d), and Figure \ref{fig:10}(d), respectively. It is shown that intermediate views can bridge the gap between aerial and ground images and increase their feature matching performance. Noticeably, the dimension of rendered images is almost the same as the image size used in SuperPoint for feature detection. In conclusion, the proposed algorithm can obviously increase feature matching of aerial and ground images.

\begin{figure}[htbp]
    \centering
    \subfloat[]{\includegraphics{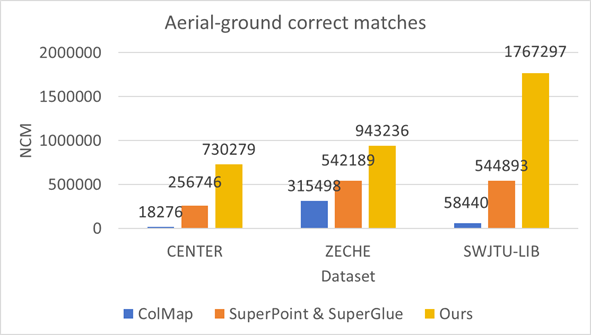}}
    \\
    \subfloat[]{\includegraphics{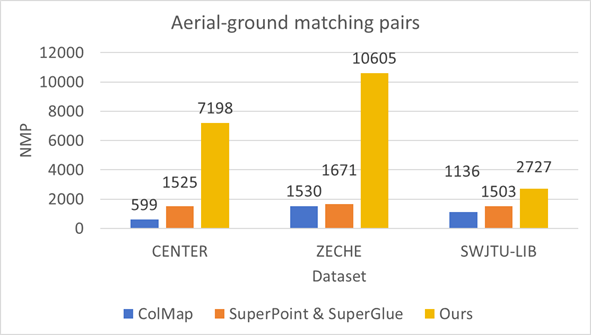}}
    \caption{The statistic of aerial-ground feature matching results in terms of NCM and NMP.}
    \label{fig:11}
\end{figure}

To evaluate the overall performance of the proposed algorithm, feature matching is then conducted for aerial-ground image pairs of these three datasets. Because of the large viewpoint changes, it is hard to retrieval correct match pairs based on image retrieval. In this evaluation, image pairs are selected by using an exhaustive way. In addition, two metrics NCM and NMP are used for performance comparison. The former indicates the number of correct matches after RANSAN-based outlier removal; the latter represents the number of match pairs that have at least 15 correct matches, which would be used to guide subsequent ISfM reconstruction.

Figure \ref{fig:11} presents the statistical results of aerial-ground feature matching for the three datasets. It is clearly shown that for these two metrics, the proposed algorithm achieves better performance when compared with other two methods. For the metric NCM, the increase ratio of the proposed algorithm is 3.72, 5.35, and 0.81 for the three datasets, respectively, compared with SuperPoint \& SuperGlue. It indicates that many more matches can be found as shown in Figure \ref{fig:11}(a), which further leads to more validated match pairs in Figure \ref{fig:11}(b). Therefore, the proposed algorithm can provide reliable matches for aerial and ground images.

\subsection{Application of matching result in 3DGS rendering}
\label{sec:4.5}
\begin{figure}[htbp]
\centering
    \subfloat[CENTER]{\includegraphics[width = 0.33\textwidth]{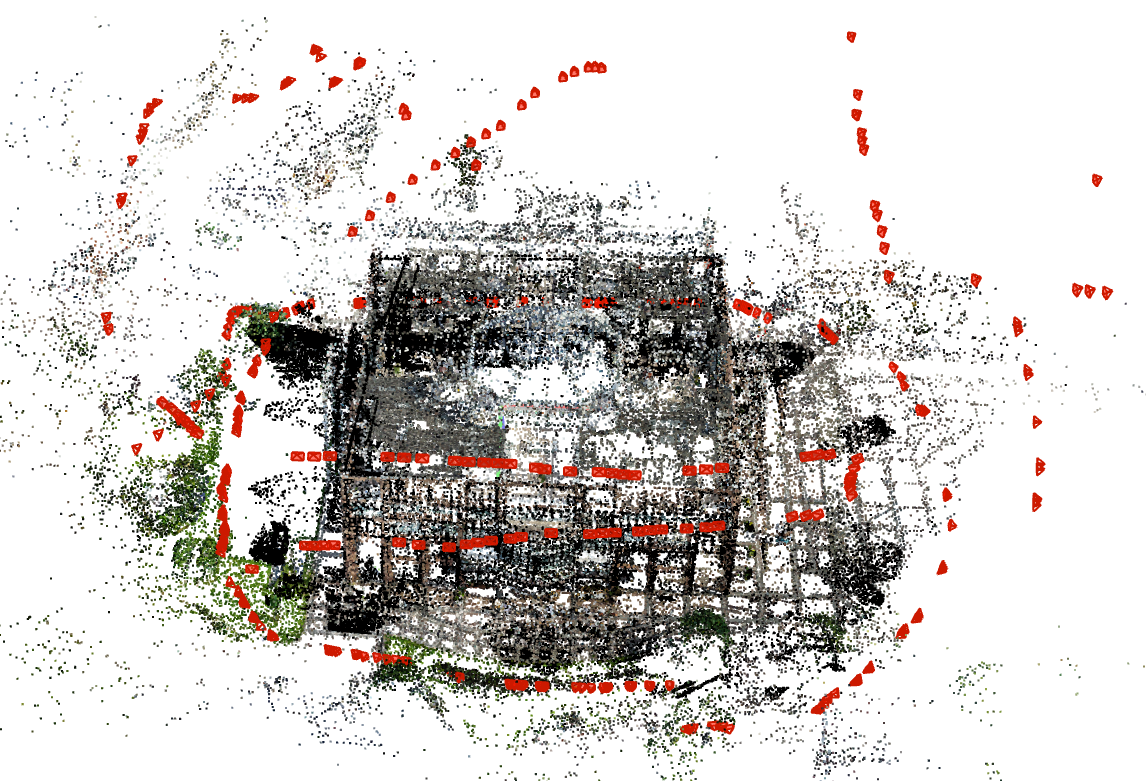}}
    \hfill
    \subfloat[ZECHE]{\includegraphics[width = 0.33\textwidth]{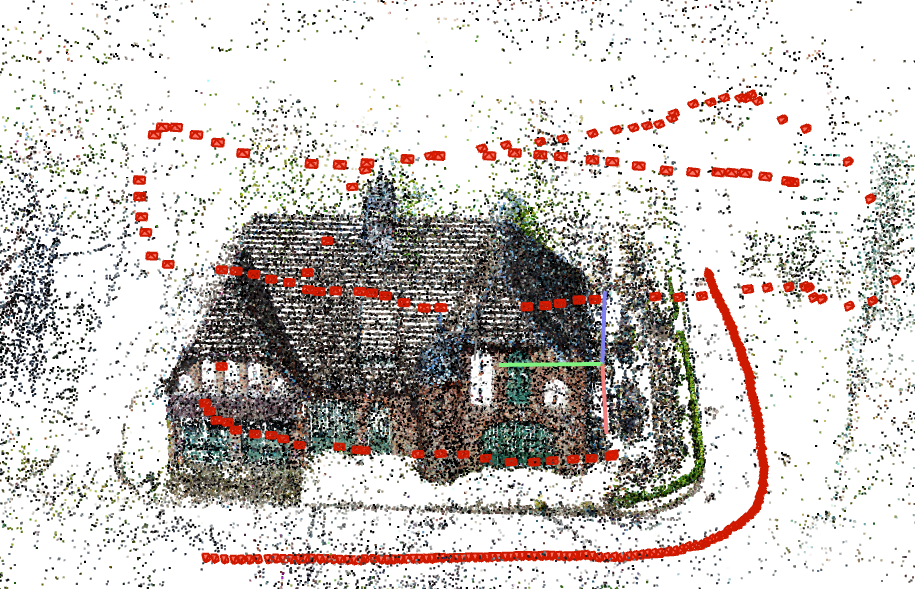}}
    \hfill
    \subfloat[SWJTU-LIB]{\includegraphics[width = 0.33\textwidth]{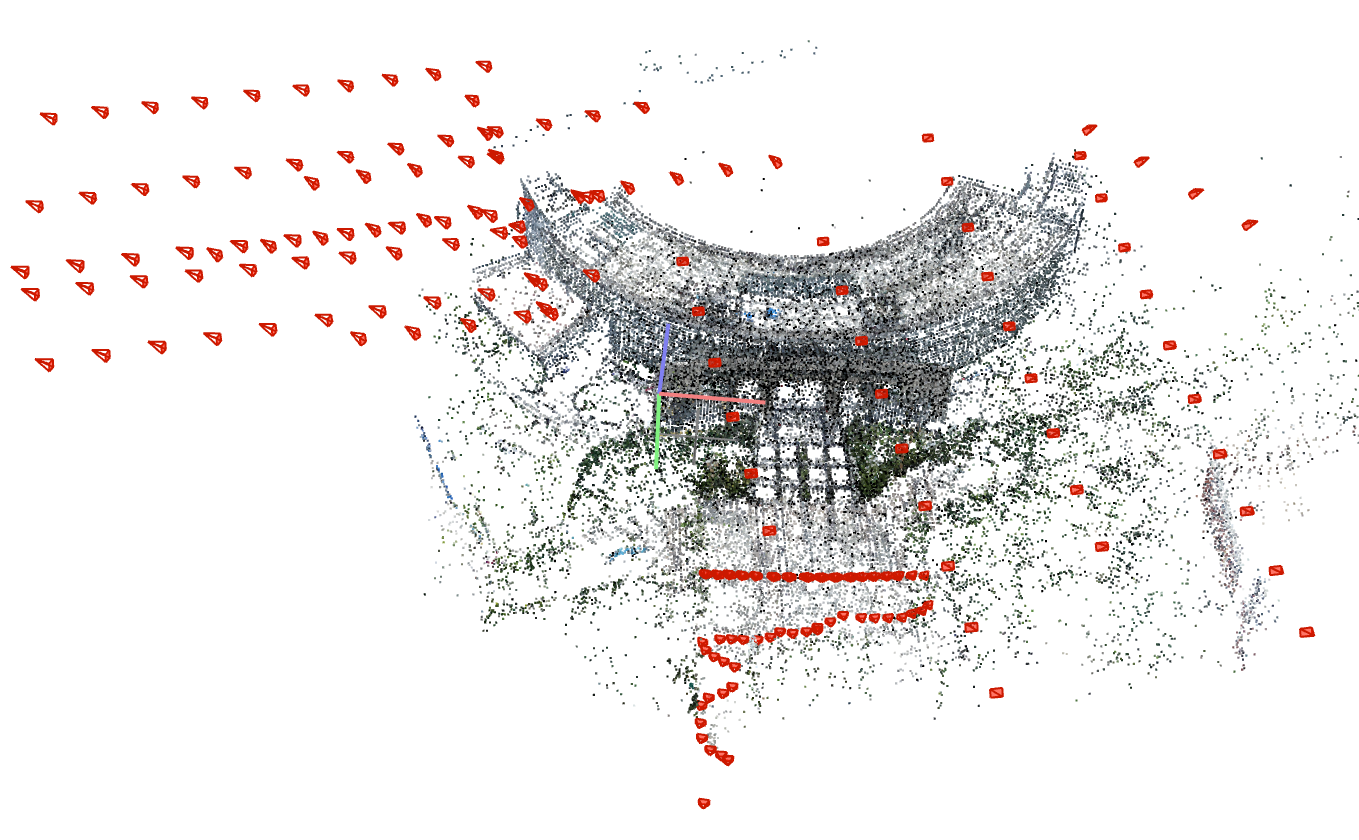}}
    \caption{SfM reconstruction of the three datasets. The red rectangles indicate oriented images, and 3D points are rendered by image colors.}
    \label{fig:12}
\end{figure}

ISfM reconstruction can be executed to create sparse models for scene rendering based on 3D Gaussian Splatting using feature matching results of aerial and ground images. The input of 3D Gaussian Splatting is a sparse model with point clouds for Gaussian initialization and oriented images for scene training. In this section, ISfM is utilized to resume camera poses and 3D points. Table \ref{tab:6} presents the statistical results of sparse reconstruction in the terms of NRI and N3P, which indicates the number of registered images and resumed 3D points, respectively. We can see that by using the matching results, all images in three datasets can be successfully registered to create entire sparse models, as shown in Figure \ref{fig:12}.

\begin{table}[htbp]
    \centering
    \caption{Results for the three datasets on the sparse reconstruction.}
    \begin{tabular}{crrr}
    \hline
    \multirow{2}{*}{\textbf{Dataset}} & \multicolumn{2}{c}{\textbf{NRI}}  & \multirow{2}{*}{\textbf{N3P}} \\ \cline{2-3}
                                      & \textbf{Aerial} & \textbf{Ground} &                               \\ \hline
    CENTER                            & 146/146         & 203/203         & 245,623                       \\
    ZECHE                             & 172/172         & 147/147         & 324,683                       \\
    SWJTU-LIB                         & 123/123         & 78/78           & 142,408                       \\ \hline
    \end{tabular}
    \label{tab:6}
\end{table}

\begin{figure}[htbp]    
\centering
    \hspace{0.3cm} Aerial images only \hspace{3cm} Aerial-ground images   
    \subfloat[CENTER]{\includegraphics[width = \textwidth]{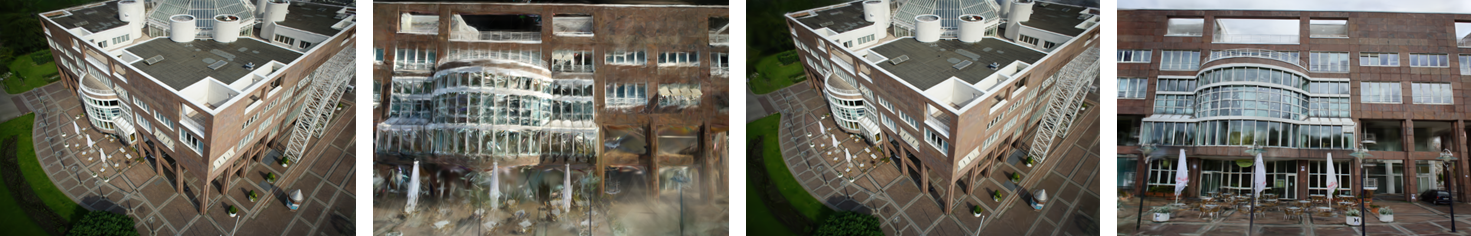}}
    \\
    \subfloat[ZECHE]{\includegraphics[width = \textwidth]{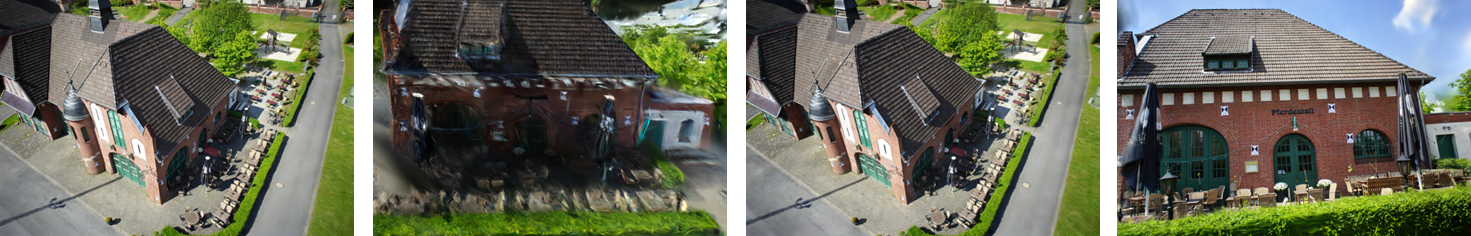}}
    \\
    \subfloat[SWJTU-LIB]{\includegraphics[width = \textwidth]{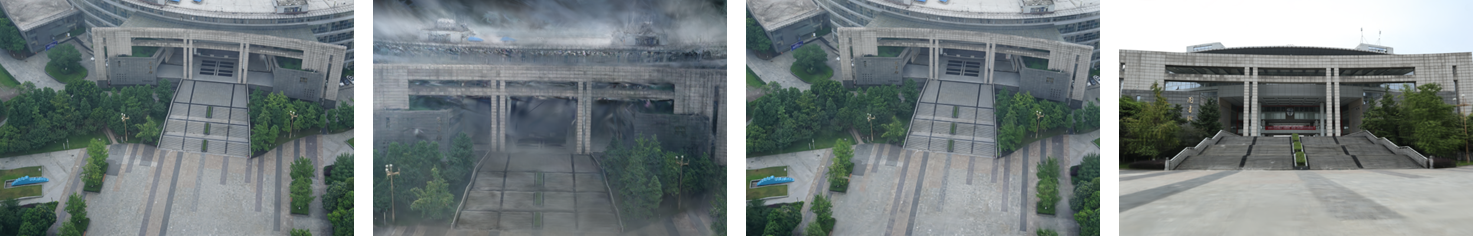}}
    \caption{The illustration of 3DGS scene rendering with varying configurations.}
    \label{fig:13}
\end{figure}

By using the reconstructed sparse models, 3D Gaussian Splatting can be utilized to render photorealistic scenes. Compared with classical SfM-MVS based pipeline, 3D Gaussian Splatting can render scenes with high precision and efficiency. To compare the effect of integrating aerial and ground images, two individual tests are conducted in this section. The first one uses aerial images only for scene rendering; the second one uses both aerial and ground images for scene rendering. Both time cost and render quality are evaluated in this section. Table \ref{tab:7} lists the time cost of scene rendering for two configurations. It is clearly shown that less time costs are consumed by the configuration with aerial-ground images when compared with that with aerial images only, which is 14.0 min, 23.0 min, and 21.0 min for the three datasets. The main reason is that by using both aerial and ground images, 3D Gaussian Splatting can converge rapidly with required precision, especially for the near-ground regions that can not be observed enough from aerial images. Figure \ref{fig:13} compares the render effects from the scene generated with varying configurations. We can see that photorealistic scenes can be created by integrating both aerial and ground images based on 3D Gaussian Splatting. Thus, the proposed algorithm can provide reliable feature matches to achieve accurate and complete scene rendering.

\begin{table}[htbp]
    \centering
    \caption{The time cost of scene rendering for two configurations (unit in minutes).}
    \begin{tabular}{crr}
    \hline
    \textbf{Dataset} & \textbf{Aerial images only} & \textbf{Aerial-ground images} \\ \hline
    CENTER           & 20.0                        & 14.0                          \\
    ZECHE            & 25.0                        & 23.0                          \\
    SWJTU-LIB        & 28.0                        & 21.0                          \\ \hline
    \end{tabular}
    \label{tab:7}
\end{table}

\begin{figure}[htbp]    
\centering
    MVS \hspace{6cm} 3DGS   
    \subfloat[CENTER]{\includegraphics[width = \textwidth]{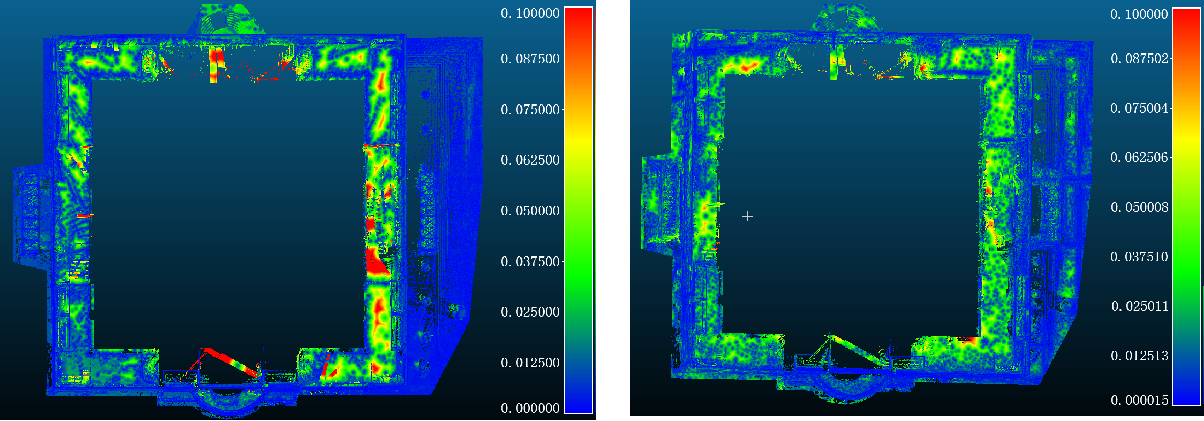}}
    \\
    \subfloat[ZECHE]{\includegraphics[width = \textwidth]{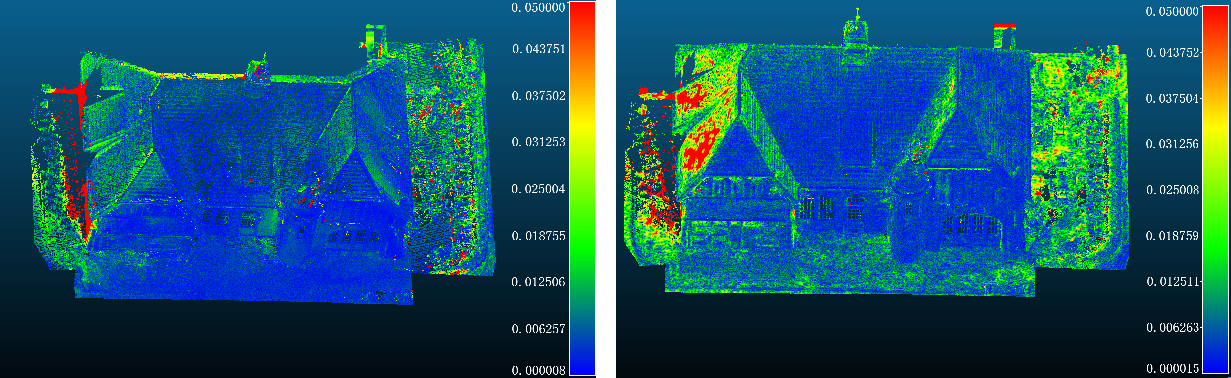}}
    \caption{The illustration of cloud-to-cloud comparisons for both MVS and 3DGS methods with respect to the ground truth data (unit in cm).}
    \label{fig:add2}
\end{figure}

In addition to rendering, we also evaluated the geometric accuracy. Using cloud-to-cloud comparison, we assessed the relative Euclidean distance between the generated point clouds and the reference data, providing a quantitative measure of the accuracy of different reconstruction pipelines. Considering that the CENTER and ZECHE datasets contain LiDAR data, we used them as the ground truth data (GT), while the SWJTU-LIB dataset does not include LiDAR data and thus could not be evaluated. For each dataset, we generated dense point clouds using the SfM-MVS workflow (ColMap), as well as the previously obtained 3DGS point clouds, and then registered them to the GT (Figure \ref{fig:add2}). Finally, as shown in Table \ref{tab:add2}, we extracted metrics. In the illustrations, the left side shows the results from the MVS pipeline, and the right side shows those from the 3DGS pipeline. The color scale indicates the distance metric, with warmer colors representing greater distances from the reference data. The table lists the quantitative results, including the maximum distance, mean distance, and standard deviation for each dataset and pipeline. These metrics are crucial for assessing the precision and consistency of the reconstruction methods.

\begin{table}[htbp]
    \centering
    \caption{Metrics of cloud-to-cloud comparisons (unit in cm).}
    \begin{tabular}{lllll}
    \hline
    \textbf{Dataset}        & \textbf{Pipeline} & \textbf{Max Distance} & \textbf{Mean distance} & \textbf{Standard deviation} \\ \hline
    \multirow{2}{*}{CENTER} & MVS               & 29.66                 & 0.76                   & 0.96                        \\
                            & 3DGS              & 11.39                 & 1.12                   & 0.88                        \\
    \multirow{2}{*}{ZECHE}  & MVS               & 46.45                 & 0.63                   & 1.12                        \\
                            & 3DGS              & 16.58                 & 0.78                   & 0.67                        \\ \hline
    \end{tabular}
    \label{tab:add2}
\end{table}

Specifically, Figure \ref{fig:add2}, the visualization  from the MVS pipeline shows more blue areas, indicating smaller distances, and the Table \ref{tab:add2} also shows that the MVS mean distances are 0.76 cm and 0.63 cm, lower than 3DGS's 1.12 cm and 0.78 cm. This is because the original 3DGS only constrains color, so it is not as geometrically accurate as MVS overall. However, 3DGS has a clear advantage in the maximum distance and standard deviation metrics, with the MVS pipeline's visualization showing very noticeable red areas. For the CENTER dataset, the MVS pipeline's maximum distance is 29.66 cm, with a standard deviation of 0.96; while the 3DGS pipeline reduces this to 11.39 cm and 0.88. On the ZECHE dataset, the MVS pipeline's maximum distance is 46.45 cm, with a standard deviation of 1.12; in contrast, the 3DGS pipeline significantly reduces this to 16.58 cm and 0.67. This indicates that the 3DGS method is more robust when dealing with datasets that have complex structures or varying lighting conditions, and in the future, geometric constraints can be added to 3DGS to achieve more accurate 3D modeling.

\section{Discussions}
\label{sec:5}
(1) Advantages of Integrating Aerial and Ground Images

Compared with using aerial images alone, the integration of aerial and ground images offers advantages in computational time, reconstruction accuracy, and completeness of scene coverage. Aerial images provide broad coverage but lack detail, performing well for flat areas yet poorly for vertical surfaces. Ground images supply high-resolution details, especially for vertical structures, compensating for the deficiencies of aerial images. This combination enables high-precision, all-around reconstruction while reducing costs and improving data acquisition efficiency in small areas.

(2) Challenges of 3D Gaussian Splatting in Large-Scale Reconstructions

Though efficient, 3D Gaussian Splatting faces challenges in large-scale reconstructions due to computational complexity and memory usage. Large scenes require numerous Gaussian points for detail representation, constrained by GPU memory. Global optimization demands extensive iterations, which are time-consuming and prone to instability. Complex lighting and image variations further complicate the process. Future research will explore strategies like block-wise reconstruction, incorporating prior knowledge, and parallel processing to address these issues.

(3) Error Sources in the Reconstruction Process

Errors in the reconstruction process are complex, especially regarding feature matching reliability and depth estimation accuracy when rendering intermediate views. For feature matching, intermediate views may not fully eliminate perspective differences, leading to inaccurate feature point extraction or mismatches, especially in areas with sparse or repetitive textures. Rendering quality issues can also affect feature point positioning. For depth estimation, information loss, lighting differences, and accumulated errors during rendering can result in inaccurate depth estimation.

\section{Conclusions}
\label{sec:6}
For reliable feature matching between aerial and ground images, this study proposes an intermediate view rendering based solution. Similar to view rendering-based methods, the core idea is to use intermediate views to alleviate viewpoint changes of aerial and ground images. The main workflow consists of: (1) SfM reconstruction from aerial images; (2) intermediate view rendering using 3D Gaussian Splatting; and (3) reliable feature matching with the aid of intermediate view. The validation of the proposed solution has been verified by using real datasets and compared with other widely used methods. The results demonstrate that the proposed algorithm can provide reliable feature matches for aerial and ground images and can achieve accurate and complete scene rendering for 3D Gaussian Splatting. In future work, two major tests can be further conducted. On the one hand, the improved version of 3DGS can be used and verified in aerial-ground image matching; on the other hand, mesh models can be extracted from rendered scenes for quantitative precision evaluation.

\section*{Disclosure statement}
No potential conflict of interest was reported by the author(s).

\section*{Data availability statement}
Data is not available due to ethical and legal restrictions.

\section*{Acknowledgement}
This research was funded by the National Natural Science Foundation of China (Grant No. 42371442), the Hubei Provincial Natural Science Foundation of China (Grant No. 2023AFB568).

 \bibliographystyle{elsarticle-harv} 
 \bibliography{elsarticle-harv}

\begin{thebibliography}{42}
\expandafter\ifx\csname natexlab\endcsname\relax\def\natexlab#1{#1}\fi
\providecommand{\url}[1]{\texttt{#1}}
\providecommand{\href}[2]{#2}
\providecommand{\path}[1]{#1}
\providecommand{\DOIprefix}{doi:}
\providecommand{\ArXivprefix}{arXiv:}
\providecommand{\URLprefix}{URL: }
\providecommand{\Pubmedprefix}{pmid:}
\providecommand{\doi}[1]{\href{http://dx.doi.org/#1}{\path{#1}}}
\providecommand{\Pubmed}[1]{\href{pmid:#1}{\path{#1}}}
\providecommand{\bibinfo}[2]{#2}
\ifx\xfnm\relax \def\xfnm[#1]{\unskip,\space#1}\fi
\bibitem[{Besl and McKay(1992)}]{besl1992method}
\bibinfo{author}{Besl, P.J.}, \bibinfo{author}{McKay, N.D.},
  \bibinfo{year}{1992}.
\newblock \bibinfo{title}{Method for registration of 3-d shapes}, in:
  \bibinfo{booktitle}{Sensor fusion IV: control paradigms and data structures},
  \bibinfo{organization}{Spie}. pp. \bibinfo{pages}{586--606}.
\bibitem[{Chen et~al.(2022)Chen, Luo, Zhou, Tian, Zhen, Fang, Mckinnon, Tsin
  and Quan}]{chen2022aspanformer}
\bibinfo{author}{Chen, H.}, \bibinfo{author}{Luo, Z.}, \bibinfo{author}{Zhou,
  L.}, \bibinfo{author}{Tian, Y.}, \bibinfo{author}{Zhen, M.},
  \bibinfo{author}{Fang, T.}, \bibinfo{author}{Mckinnon, D.},
  \bibinfo{author}{Tsin, Y.}, \bibinfo{author}{Quan, L.}, \bibinfo{year}{2022}.
\newblock \bibinfo{title}{Aspanformer: Detector-free image matching with
  adaptive span transformer}, in: \bibinfo{booktitle}{European Conference on
  Computer Vision}, \bibinfo{organization}{Springer}. pp.
  \bibinfo{pages}{20--36}.
\bibitem[{DeTone et~al.(2018)DeTone, Malisiewicz and
  Rabinovich}]{detone2018superpoint}
\bibinfo{author}{DeTone, D.}, \bibinfo{author}{Malisiewicz, T.},
  \bibinfo{author}{Rabinovich, A.}, \bibinfo{year}{2018}.
\newblock \bibinfo{title}{Superpoint: Self-supervised interest point detection
  and description}, in: \bibinfo{booktitle}{Proceedings of the IEEE conference
  on computer vision and pattern recognition workshops}, pp.
  \bibinfo{pages}{224--236}.
\bibitem[{Dusmanu et~al.(2019)Dusmanu, Rocco, Pajdla, Pollefeys, Sivic, Torii
  and Sattler}]{dusmanu2019d2}
\bibinfo{author}{Dusmanu, M.}, \bibinfo{author}{Rocco, I.},
  \bibinfo{author}{Pajdla, T.}, \bibinfo{author}{Pollefeys, M.},
  \bibinfo{author}{Sivic, J.}, \bibinfo{author}{Torii, A.},
  \bibinfo{author}{Sattler, T.}, \bibinfo{year}{2019}.
\newblock \bibinfo{title}{D2-net: A trainable cnn for joint description and
  detection of local features}, in: \bibinfo{booktitle}{Proceedings of the
  ieee/cvf conference on computer vision and pattern recognition}, pp.
  \bibinfo{pages}{8092--8101}.
\bibitem[{Edstedt et~al.(2024)Edstedt, Sun, B{\"o}kman, Wadenb{\"a}ck and
  Felsberg}]{edstedt2024roma}
\bibinfo{author}{Edstedt, J.}, \bibinfo{author}{Sun, Q.},
  \bibinfo{author}{B{\"o}kman, G.}, \bibinfo{author}{Wadenb{\"a}ck, M.},
  \bibinfo{author}{Felsberg, M.}, \bibinfo{year}{2024}.
\newblock \bibinfo{title}{Roma: Robust dense feature matching}, in:
  \bibinfo{booktitle}{Proceedings of the IEEE/CVF Conference on Computer Vision
  and Pattern Recognition}, pp. \bibinfo{pages}{19790--19800}.
\bibitem[{Fu et~al.(2024)Fu, Liu, Kulkarni, Kautz, Efros and
  Wang}]{fu2024colmap}
\bibinfo{author}{Fu, Y.}, \bibinfo{author}{Liu, S.}, \bibinfo{author}{Kulkarni,
  A.}, \bibinfo{author}{Kautz, J.}, \bibinfo{author}{Efros, A.A.},
  \bibinfo{author}{Wang, X.}, \bibinfo{year}{2024}.
\newblock \bibinfo{title}{Colmap-free 3d gaussian splatting}, in:
  \bibinfo{booktitle}{Proceedings of the IEEE/CVF Conference on Computer Vision
  and Pattern Recognition}, pp. \bibinfo{pages}{20796--20805}.
\bibitem[{Gao et~al.(2018a)Gao, Hu, Cui, Shen and Hu}]{gao2018accurate}
\bibinfo{author}{Gao, X.}, \bibinfo{author}{Hu, L.}, \bibinfo{author}{Cui, H.},
  \bibinfo{author}{Shen, S.}, \bibinfo{author}{Hu, Z.}, \bibinfo{year}{2018}a.
\newblock \bibinfo{title}{Accurate and efficient ground-to-aerial model
  alignment}.
\newblock \bibinfo{journal}{Pattern Recognition} \bibinfo{volume}{76},
  \bibinfo{pages}{288--302}.
\bibitem[{Gao et~al.(2019)Gao, Shen, Hu and Wang}]{gao2019ground}
\bibinfo{author}{Gao, X.}, \bibinfo{author}{Shen, S.}, \bibinfo{author}{Hu,
  Z.}, \bibinfo{author}{Wang, Z.}, \bibinfo{year}{2019}.
\newblock \bibinfo{title}{Ground and aerial meta-data integration for
  localization and reconstruction: A review}.
\newblock \bibinfo{journal}{Pattern Recognition Letters} \bibinfo{volume}{127},
  \bibinfo{pages}{202--214}.
\bibitem[{Gao et~al.(2018b)Gao, Shen, Zhou, Cui, Zhu and Hu}]{gao2018ancient}
\bibinfo{author}{Gao, X.}, \bibinfo{author}{Shen, S.}, \bibinfo{author}{Zhou,
  Y.}, \bibinfo{author}{Cui, H.}, \bibinfo{author}{Zhu, L.},
  \bibinfo{author}{Hu, Z.}, \bibinfo{year}{2018}b.
\newblock \bibinfo{title}{Ancient chinese architecture 3d preservation by
  merging ground and aerial point clouds}.
\newblock \bibinfo{journal}{ISPRS Journal of Photogrammetry and Remote Sensing}
  \bibinfo{volume}{143}, \bibinfo{pages}{72--84}.
\bibitem[{Ge et~al.(2024)Ge, Guo, Xu, Liu, Jiang and Peng}]{ge2024rapid}
\bibinfo{author}{Ge, Y.}, \bibinfo{author}{Guo, B.}, \bibinfo{author}{Xu, G.},
  \bibinfo{author}{Liu, Y.}, \bibinfo{author}{Jiang, X.},
  \bibinfo{author}{Peng, Z.}, \bibinfo{year}{2024}.
\newblock \bibinfo{title}{Rapid 3d modelling: Clustering method based on
  dynamic load balancing strategy}.
\newblock \bibinfo{journal}{The Photogrammetric Record} \bibinfo{volume}{39},
  \bibinfo{pages}{67--86}.
\bibitem[{Hong et~al.(2024)Hong, Jung, Shin, Han, Yang, Luo and
  Kim}]{hong2024pf3plat}
\bibinfo{author}{Hong, S.}, \bibinfo{author}{Jung, J.}, \bibinfo{author}{Shin,
  H.}, \bibinfo{author}{Han, J.}, \bibinfo{author}{Yang, J.},
  \bibinfo{author}{Luo, C.}, \bibinfo{author}{Kim, S.}, \bibinfo{year}{2024}.
\newblock \bibinfo{title}{Pf3plat: Pose-free feed-forward 3d gaussian
  splatting}.
\newblock \bibinfo{journal}{arXiv preprint arXiv:2410.22128} .
\bibitem[{Hu et~al.(2015)Hu, Zhu, Du, Zhang and Ding}]{hu2015reliable}
\bibinfo{author}{Hu, H.}, \bibinfo{author}{Zhu, Q.}, \bibinfo{author}{Du, Z.},
  \bibinfo{author}{Zhang, Y.}, \bibinfo{author}{Ding, Y.},
  \bibinfo{year}{2015}.
\newblock \bibinfo{title}{Reliable spatial relationship constrained feature
  point matching of oblique aerial images}.
\newblock \bibinfo{journal}{Photogrammetric Engineering \& Remote Sensing}
  \bibinfo{volume}{81}, \bibinfo{pages}{49--58}.
\bibitem[{Jhan et~al.(2021)Jhan, Kerle and Rau}]{jhan2021integrating}
\bibinfo{author}{Jhan, J.P.}, \bibinfo{author}{Kerle, N.},
  \bibinfo{author}{Rau, J.Y.}, \bibinfo{year}{2021}.
\newblock \bibinfo{title}{Integrating uav and ground panoramic images for point
  cloud analysis of damaged building}.
\newblock \bibinfo{journal}{IEEE geoscience and remote sensing letters}
  \bibinfo{volume}{19}, \bibinfo{pages}{1--5}.
\bibitem[{Ji et~al.(2023)Ji, Zeng, Zhang and Duan}]{ji2023evaluation}
\bibinfo{author}{Ji, S.}, \bibinfo{author}{Zeng, C.}, \bibinfo{author}{Zhang,
  Y.}, \bibinfo{author}{Duan, Y.}, \bibinfo{year}{2023}.
\newblock \bibinfo{title}{An evaluation of conventional and deep learning-based
  image-matching methods on diverse datasets}.
\newblock \bibinfo{journal}{The Photogrammetric Record} \bibinfo{volume}{38},
  \bibinfo{pages}{137--159}.
\bibitem[{Jiang et~al.(2020)Jiang, Jiang and Jiang}]{jiang2020efficient}
\bibinfo{author}{Jiang, S.}, \bibinfo{author}{Jiang, C.},
  \bibinfo{author}{Jiang, W.}, \bibinfo{year}{2020}.
\newblock \bibinfo{title}{Efficient structure from motion for large-scale uav
  images: A review and a comparison of sfm tools}.
\newblock \bibinfo{journal}{ISPRS Journal of Photogrammetry and Remote Sensing}
  \bibinfo{volume}{167}, \bibinfo{pages}{230--251}.
\bibitem[{Jiang and Jiang(2017)}]{jiang2017board}
\bibinfo{author}{Jiang, S.}, \bibinfo{author}{Jiang, W.}, \bibinfo{year}{2017}.
\newblock \bibinfo{title}{On-board gnss/imu assisted feature extraction and
  matching for oblique uav images}.
\newblock \bibinfo{journal}{Remote Sensing} \bibinfo{volume}{9},
  \bibinfo{pages}{813}.
\bibitem[{Jiang et~al.(2021)Jiang, Jiang, Guo, Li and Wang}]{jiang2021learned}
\bibinfo{author}{Jiang, S.}, \bibinfo{author}{Jiang, W.}, \bibinfo{author}{Guo,
  B.}, \bibinfo{author}{Li, L.}, \bibinfo{author}{Wang, L.},
  \bibinfo{year}{2021}.
\newblock \bibinfo{title}{Learned local features for structure from motion of
  uav images: A comparative evaluation}.
\newblock \bibinfo{journal}{IEEE Journal of Selected Topics in Applied Earth
  Observations and Remote Sensing} \bibinfo{volume}{14},
  \bibinfo{pages}{10583--10597}.
\bibitem[{Kerbl et~al.(2023)Kerbl, Kopanas, Leimk{\"u}hler and
  Drettakis}]{kerbl20233d}
\bibinfo{author}{Kerbl, B.}, \bibinfo{author}{Kopanas, G.},
  \bibinfo{author}{Leimk{\"u}hler, T.}, \bibinfo{author}{Drettakis, G.},
  \bibinfo{year}{2023}.
\newblock \bibinfo{title}{3d gaussian splatting for real-time radiance field
  rendering.}
\newblock \bibinfo{journal}{ACM Trans. Graph.} \bibinfo{volume}{42},
  \bibinfo{pages}{139--1}.
\bibitem[{Li et~al.(2023a)Li, Huang, Yu and Jiang}]{li2023optimized}
\bibinfo{author}{Li, Q.}, \bibinfo{author}{Huang, H.}, \bibinfo{author}{Yu,
  W.}, \bibinfo{author}{Jiang, S.}, \bibinfo{year}{2023}a.
\newblock \bibinfo{title}{Optimized views photogrammetry: Precision analysis
  and a large-scale case study in qingdao}.
\newblock \bibinfo{journal}{IEEE Journal of Selected Topics in Applied Earth
  Observations and Remote Sensing} \bibinfo{volume}{16},
  \bibinfo{pages}{1144--1159}.
\bibitem[{Li et~al.(2023b)Li, Wu, Li and Chen}]{li2023fusion}
\bibinfo{author}{Li, Z.}, \bibinfo{author}{Wu, B.}, \bibinfo{author}{Li, Y.},
  \bibinfo{author}{Chen, Z.}, \bibinfo{year}{2023}b.
\newblock \bibinfo{title}{Fusion of aerial, mms and backpack images and point
  clouds for optimized 3d mapping in urban areas}.
\newblock \bibinfo{journal}{ISPRS Journal of Photogrammetry and Remote Sensing}
  \bibinfo{volume}{202}, \bibinfo{pages}{463--478}.
\bibitem[{Liang et~al.(2023)Liang, Yang, Mu and Cui}]{liang2023robust}
\bibinfo{author}{Liang, Y.}, \bibinfo{author}{Yang, Y.}, \bibinfo{author}{Mu,
  Y.}, \bibinfo{author}{Cui, T.}, \bibinfo{year}{2023}.
\newblock \bibinfo{title}{Robust fusion of multi-source images for accurate 3d
  reconstruction of complex urban scenes}.
\newblock \bibinfo{journal}{Remote Sensing} \bibinfo{volume}{15},
  \bibinfo{pages}{5302}.
\bibitem[{Liu et~al.(2023)Liu, Yin, Liu and Lu}]{liu2023tie}
\bibinfo{author}{Liu, J.}, \bibinfo{author}{Yin, H.}, \bibinfo{author}{Liu,
  B.}, \bibinfo{author}{Lu, P.}, \bibinfo{year}{2023}.
\newblock \bibinfo{title}{Tie point matching between terrestrial and aerial
  images based on patch variational refinement}.
\newblock \bibinfo{journal}{Remote Sensing} \bibinfo{volume}{15},
  \bibinfo{pages}{968}.
\bibitem[{Luo et~al.(2019)Luo, Shen, Zhou, Zhang, Yao, Li, Fang and
  Quan}]{luo2019contextdesc}
\bibinfo{author}{Luo, Z.}, \bibinfo{author}{Shen, T.}, \bibinfo{author}{Zhou,
  L.}, \bibinfo{author}{Zhang, J.}, \bibinfo{author}{Yao, Y.},
  \bibinfo{author}{Li, S.}, \bibinfo{author}{Fang, T.}, \bibinfo{author}{Quan,
  L.}, \bibinfo{year}{2019}.
\newblock \bibinfo{title}{Contextdesc: Local descriptor augmentation with
  cross-modality context}, in: \bibinfo{booktitle}{Proceedings of the IEEE/CVF
  conference on computer vision and pattern recognition}, pp.
  \bibinfo{pages}{2527--2536}.
\bibitem[{Luo et~al.(2018)Luo, Shen, Zhou, Zhu, Zhang, Yao, Fang and
  Quan}]{luo2018geodesc}
\bibinfo{author}{Luo, Z.}, \bibinfo{author}{Shen, T.}, \bibinfo{author}{Zhou,
  L.}, \bibinfo{author}{Zhu, S.}, \bibinfo{author}{Zhang, R.},
  \bibinfo{author}{Yao, Y.}, \bibinfo{author}{Fang, T.}, \bibinfo{author}{Quan,
  L.}, \bibinfo{year}{2018}.
\newblock \bibinfo{title}{Geodesc: Learning local descriptors by integrating
  geometry constraints}, in: \bibinfo{booktitle}{Proceedings of the European
  conference on computer vision (ECCV)}, pp. \bibinfo{pages}{168--183}.
\bibitem[{Luo et~al.(2020)Luo, Zhou, Bai, Chen, Zhang, Yao, Li, Fang and
  Quan}]{luo2020aslfeat}
\bibinfo{author}{Luo, Z.}, \bibinfo{author}{Zhou, L.}, \bibinfo{author}{Bai,
  X.}, \bibinfo{author}{Chen, H.}, \bibinfo{author}{Zhang, J.},
  \bibinfo{author}{Yao, Y.}, \bibinfo{author}{Li, S.}, \bibinfo{author}{Fang,
  T.}, \bibinfo{author}{Quan, L.}, \bibinfo{year}{2020}.
\newblock \bibinfo{title}{Aslfeat: Learning local features of accurate shape
  and localization}, in: \bibinfo{booktitle}{Proceedings of the IEEE/CVF
  conference on computer vision and pattern recognition}, pp.
  \bibinfo{pages}{6589--6598}.
\bibitem[{Mildenhall et~al.(2021)Mildenhall, Srinivasan, Tancik, Barron,
  Ramamoorthi and Ng}]{mildenhall2021nerf}
\bibinfo{author}{Mildenhall, B.}, \bibinfo{author}{Srinivasan, P.P.},
  \bibinfo{author}{Tancik, M.}, \bibinfo{author}{Barron, J.T.},
  \bibinfo{author}{Ramamoorthi, R.}, \bibinfo{author}{Ng, R.},
  \bibinfo{year}{2021}.
\newblock \bibinfo{title}{Nerf: Representing scenes as neural radiance fields
  for view synthesis}.
\newblock \bibinfo{journal}{Communications of the ACM} \bibinfo{volume}{65},
  \bibinfo{pages}{99--106}.
\bibitem[{Mishchuk et~al.(2017)Mishchuk, Mishkin, Radenovic and
  Matas}]{mishchuk2017working}
\bibinfo{author}{Mishchuk, A.}, \bibinfo{author}{Mishkin, D.},
  \bibinfo{author}{Radenovic, F.}, \bibinfo{author}{Matas, J.},
  \bibinfo{year}{2017}.
\newblock \bibinfo{title}{Working hard to know your neighbor's margins: Local
  descriptor learning loss}.
\newblock \bibinfo{journal}{Advances in neural information processing systems}
  \bibinfo{volume}{30}.
\bibitem[{Nex et~al.(2015)Nex, Gerke, Remondino, Przybilla, B{\"a}umker and
  Zurhorst}]{nex2015isprs}
\bibinfo{author}{Nex, F.}, \bibinfo{author}{Gerke, M.},
  \bibinfo{author}{Remondino, F.}, \bibinfo{author}{Przybilla, H.J.},
  \bibinfo{author}{B{\"a}umker, M.}, \bibinfo{author}{Zurhorst, A.},
  \bibinfo{year}{2015}.
\newblock \bibinfo{title}{Isprs benchmark for multi-platform photogrammetry}.
\newblock \bibinfo{journal}{ISPRS Annals of the Photogrammetry, Remote Sensing
  and Spatial Information Sciences} \bibinfo{volume}{2},
  \bibinfo{pages}{135--142}.
\bibitem[{Sarlin et~al.(2020)Sarlin, DeTone, Malisiewicz and
  Rabinovich}]{sarlin2020superglue}
\bibinfo{author}{Sarlin, P.E.}, \bibinfo{author}{DeTone, D.},
  \bibinfo{author}{Malisiewicz, T.}, \bibinfo{author}{Rabinovich, A.},
  \bibinfo{year}{2020}.
\newblock \bibinfo{title}{Superglue: Learning feature matching with graph
  neural networks}, in: \bibinfo{booktitle}{Proceedings of the IEEE/CVF
  conference on computer vision and pattern recognition}, pp.
  \bibinfo{pages}{4938--4947}.
\bibitem[{Schonberger and Frahm(2016)}]{schonberger2016structure}
\bibinfo{author}{Schonberger, J.L.}, \bibinfo{author}{Frahm, J.M.},
  \bibinfo{year}{2016}.
\newblock \bibinfo{title}{Structure-from-motion revisited}, in:
  \bibinfo{booktitle}{Proceedings of the IEEE conference on computer vision and
  pattern recognition}, pp. \bibinfo{pages}{4104--4113}.
\bibitem[{Shan et~al.(2014)Shan, Wu, Curless, Furukawa, Hernandez and
  Seitz}]{shan2014accurate}
\bibinfo{author}{Shan, Q.}, \bibinfo{author}{Wu, C.}, \bibinfo{author}{Curless,
  B.}, \bibinfo{author}{Furukawa, Y.}, \bibinfo{author}{Hernandez, C.},
  \bibinfo{author}{Seitz, S.M.}, \bibinfo{year}{2014}.
\newblock \bibinfo{title}{Accurate geo-registration by ground-to-aerial image
  matching}, in: \bibinfo{booktitle}{2014 2nd International Conference on 3D
  Vision}, \bibinfo{organization}{IEEE}. pp. \bibinfo{pages}{525--532}.
\bibitem[{Song et~al.(2019)Song, Jung, Gwak and Lee}]{song2019oblique}
\bibinfo{author}{Song, W.H.}, \bibinfo{author}{Jung, H.G.},
  \bibinfo{author}{Gwak, I.Y.}, \bibinfo{author}{Lee, S.W.},
  \bibinfo{year}{2019}.
\newblock \bibinfo{title}{Oblique aerial image matching based on iterative
  simulation and homography evaluation}.
\newblock \bibinfo{journal}{Pattern Recognition} \bibinfo{volume}{87},
  \bibinfo{pages}{317--331}.
\bibitem[{Sun et~al.(2021)Sun, Shen, Wang, Bao and Zhou}]{sun2021loftr}
\bibinfo{author}{Sun, J.}, \bibinfo{author}{Shen, Z.}, \bibinfo{author}{Wang,
  Y.}, \bibinfo{author}{Bao, H.}, \bibinfo{author}{Zhou, X.},
  \bibinfo{year}{2021}.
\newblock \bibinfo{title}{Loftr: Detector-free local feature matching with
  transformers}, in: \bibinfo{booktitle}{Proceedings of the IEEE/CVF conference
  on computer vision and pattern recognition}, pp. \bibinfo{pages}{8922--8931}.
\bibitem[{Tian et~al.(2017)Tian, Fan and Wu}]{tian2017l2}
\bibinfo{author}{Tian, Y.}, \bibinfo{author}{Fan, B.}, \bibinfo{author}{Wu,
  F.}, \bibinfo{year}{2017}.
\newblock \bibinfo{title}{L2-net: Deep learning of discriminative patch
  descriptor in euclidean space}, in: \bibinfo{booktitle}{Proceedings of the
  IEEE conference on computer vision and pattern recognition}, pp.
  \bibinfo{pages}{661--669}.
\bibitem[{Wang et~al.(2022)Wang, Zhang, Yang, Peng and
  Stiefelhagen}]{wang2022matchformer}
\bibinfo{author}{Wang, Q.}, \bibinfo{author}{Zhang, J.}, \bibinfo{author}{Yang,
  K.}, \bibinfo{author}{Peng, K.}, \bibinfo{author}{Stiefelhagen, R.},
  \bibinfo{year}{2022}.
\newblock \bibinfo{title}{Matchformer: Interleaving attention in transformers
  for feature matching}, in: \bibinfo{booktitle}{Proceedings of the Asian
  Conference on Computer Vision}, pp. \bibinfo{pages}{2746--2762}.
\bibitem[{Wang et~al.(2023)Wang, Xiang, Niu, Mao, Huang and
  Zhang}]{wang2023oblique}
\bibinfo{author}{Wang, X.}, \bibinfo{author}{Xiang, H.}, \bibinfo{author}{Niu,
  W.}, \bibinfo{author}{Mao, Z.}, \bibinfo{author}{Huang, X.},
  \bibinfo{author}{Zhang, F.}, \bibinfo{year}{2023}.
\newblock \bibinfo{title}{Oblique photogrammetry supporting procedural tree
  modeling in urban areas}.
\newblock \bibinfo{journal}{ISPRS Journal of Photogrammetry and Remote Sensing}
  \bibinfo{volume}{200}, \bibinfo{pages}{120--137}.
\bibitem[{Wu et~al.(2018)Wu, Xie, Hu, Zhu and Yau}]{wu2018integration}
\bibinfo{author}{Wu, B.}, \bibinfo{author}{Xie, L.}, \bibinfo{author}{Hu, H.},
  \bibinfo{author}{Zhu, Q.}, \bibinfo{author}{Yau, E.}, \bibinfo{year}{2018}.
\newblock \bibinfo{title}{Integration of aerial oblique imagery and terrestrial
  imagery for optimized 3d modeling in urban areas}.
\newblock \bibinfo{journal}{ISPRS journal of photogrammetry and remote sensing}
  \bibinfo{volume}{139}, \bibinfo{pages}{119--132}.
\bibitem[{Wu et~al.(2008)Wu, Clipp, Li, Frahm and Pollefeys}]{wu20083d}
\bibinfo{author}{Wu, C.}, \bibinfo{author}{Clipp, B.}, \bibinfo{author}{Li,
  X.}, \bibinfo{author}{Frahm, J.M.}, \bibinfo{author}{Pollefeys, M.},
  \bibinfo{year}{2008}.
\newblock \bibinfo{title}{3d model matching with viewpoint-invariant patches
  (vip)}, in: \bibinfo{booktitle}{2008 IEEE Conference on Computer Vision and
  Pattern Recognition}, \bibinfo{organization}{IEEE}. pp.
  \bibinfo{pages}{1--8}.
\bibitem[{Xu et~al.(2024)Xu, Chen, Xu, Wang, Lu and Guo}]{xu2024local}
\bibinfo{author}{Xu, S.}, \bibinfo{author}{Chen, S.}, \bibinfo{author}{Xu, R.},
  \bibinfo{author}{Wang, C.}, \bibinfo{author}{Lu, P.}, \bibinfo{author}{Guo,
  L.}, \bibinfo{year}{2024}.
\newblock \bibinfo{title}{Local feature matching using deep learning: A
  survey}.
\newblock \bibinfo{journal}{Information Fusion} \bibinfo{volume}{107},
  \bibinfo{pages}{102344}.
\bibitem[{Ye et~al.(2024)Ye, Liu, Xu, Li, Pollefeys, Yang and Peng}]{ye2024no}
\bibinfo{author}{Ye, B.}, \bibinfo{author}{Liu, S.}, \bibinfo{author}{Xu, H.},
  \bibinfo{author}{Li, X.}, \bibinfo{author}{Pollefeys, M.},
  \bibinfo{author}{Yang, M.H.}, \bibinfo{author}{Peng, S.},
  \bibinfo{year}{2024}.
\newblock \bibinfo{title}{No pose, no problem: Surprisingly simple 3d gaussian
  splats from sparse unposed images}.
\newblock \bibinfo{journal}{arXiv preprint arXiv:2410.24207} .
\bibitem[{Zhu et~al.(2024)Zhu, Ye, Dai, Peng, Deng and Zhu}]{zhu2024vdft}
\bibinfo{author}{Zhu, B.}, \bibinfo{author}{Ye, Y.}, \bibinfo{author}{Dai, J.},
  \bibinfo{author}{Peng, T.}, \bibinfo{author}{Deng, J.}, \bibinfo{author}{Zhu,
  Q.}, \bibinfo{year}{2024}.
\newblock \bibinfo{title}{Vdft: Robust feature matching of aerial and ground
  images using viewpoint-invariant deformable feature transformation}.
\newblock \bibinfo{journal}{ISPRS Journal of Photogrammetry and Remote Sensing}
  \bibinfo{volume}{218}, \bibinfo{pages}{311--325}.
\bibitem[{Zhu et~al.(2020)Zhu, Wang, Hu, Xie, Ge and Zhang}]{zhu2020leveraging}
\bibinfo{author}{Zhu, Q.}, \bibinfo{author}{Wang, Z.}, \bibinfo{author}{Hu,
  H.}, \bibinfo{author}{Xie, L.}, \bibinfo{author}{Ge, X.},
  \bibinfo{author}{Zhang, Y.}, \bibinfo{year}{2020}.
\newblock \bibinfo{title}{Leveraging photogrammetric mesh models for
  aerial-ground feature point matching toward integrated 3d reconstruction}.
\newblock \bibinfo{journal}{ISPRS Journal of Photogrammetry and Remote Sensing}
  \bibinfo{volume}{166}, \bibinfo{pages}{26--40}.

\end{thebibliography}





\end{sloppypar}
\end{document}